\documentclass[11pt,a4paper]{article}
\usepackage[hyperref]{acl2019}
\usepackage{times}
\usepackage{latexsym}

\usepackage{url}
\usepackage[utf8]{inputenc}

\usepackage{graphicx}
\usepackage{rotating}

\usepackage{color}
\usepackage{subcaption}
\usepackage{booktabs}
\usepackage[mathscr]{eucal}
\usepackage{amsmath, amssymb}
\usepackage{caption}
\usepackage{xspace}
\usepackage{siunitx}
\usepackage{url}
\usepackage{adjustbox}
\usepackage{arydshln}
\usepackage{setspace}

\aclfinalcopy 

\title{Parallax: Visualizing and Understanding the Semantics of Embedding Spaces via Algebraic Formulae}

\author{Piero Molino \\
  Uber AI Labs \\
  San Francisco, CA, USA \\
  {\tt piero@uber.com} \\\And
  Yang Wang \\
  Uber Technologies Inc. \\
  San Francisco, CA, USA \\
  {\tt gnavvy@uber.com} \\\And
  Jiawei Zhang\thanks{\scriptsize{Work done while at Purdue University}} \\
  Facebook \\
  Menlo Park, CA, USA \\
  {\tt rivulet.zhang@gmail.com} \\
  }

\date{}

\begin{document}
\maketitle


\begin{abstract}

Embeddings are a fundamental component of many modern machine learning and natural language processing models.
Understanding them and visualizing them is essential for gathering insights about the information they capture and the behavior of the models.
In this paper, we introduce Parallax\footnote{\scriptsize{\url{http://github.com/uber-research/parallax}}}, a tool explicitly designed for this task.
Parallax allows the user to use both state-of-the-art embedding analysis methods (PCA and t-SNE) and a simple yet effective task-oriented approach where users can explicitly define the axes of the projection through algebraic formulae.
In this approach, embeddings are projected into a semantically meaningful subspace, which enhances interpretability and allows for more fine-grained analysis.
We demonstrate\footnote{\scriptsize{\url{https://youtu.be/CSkJGVsFPIg}}} the power of the tool and the proposed methodology through a series of case studies and a user study.

\end{abstract}

\section{Introduction}

Learning representations is an important part of modern machine learning and natural language processing. 
These representations are often real-valued vectors also called embeddings and are obtained both as byproducts of supervised learning or as the direct goal of unsupervised methods.
Independently of how the embeddings are learned, there is much value in understanding what information they capture, how they relate to each other and how the data they are learned from influences them.
A better understanding of the embedded space may lead to a better understanding of the data, of the problem and the behavior of the model, and may lead to critical insights in improving such models.
Because of their high-dimensional nature, they are hard to visualize effectively.

In this paper, we introduce Parallax, a tool for visualizing embedding spaces.
The most widely adopted projection techniques (Principal Component Analysis (PCA)~\citep{Pearson1901} and t-Distributed Stochastic Neighbor Embedding (t-SNE)~\citep{vanDerMaaten2008}) are available in Parallax.
They are useful for obtaining an overall view of the embedding space, but they have a few shortcomings: 1) projections may not preserve distance in the original space, 2) they are not comparable across models and 3) do not provide interpretable axes, preventing more detailed analysis and understanding.

\begin{figure}[t]
\begin{center}
\includegraphics[width=\columnwidth]{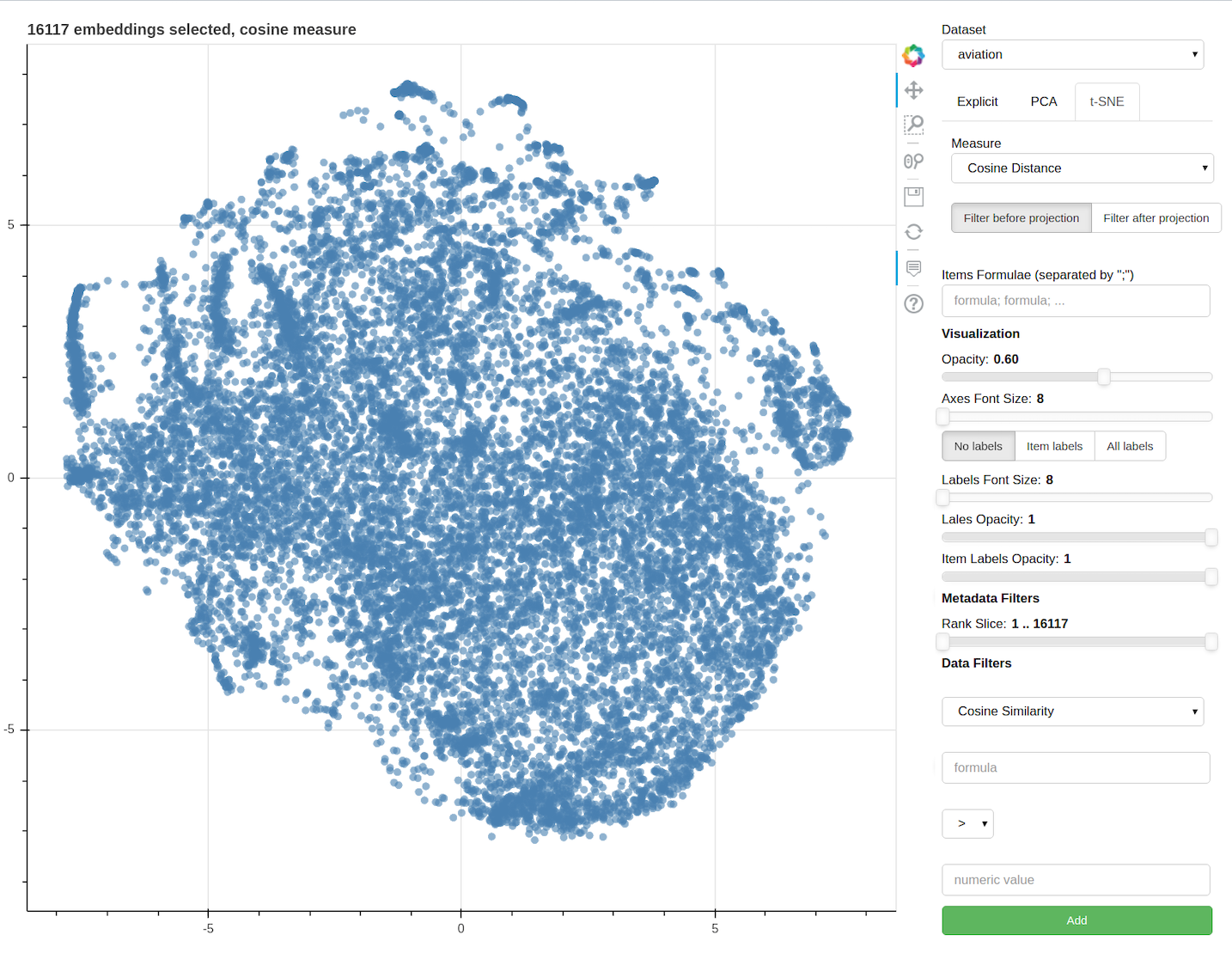}
\end{center}
  \vspace*{-1ex}
\caption{Screenshot of Parallax.}
\label{fig:parallax_screenshot}
  \vspace*{-3ex}
\end{figure}

PCA projects embeddings on a lower dimensional space that has the directions of the highest variance in the dataset as axes.
Those dimensions do not carry any interpretable meaning, so by visualizing the first two dimensions of a PCA projection, the only insight obtainable is semantic relatedness~\citep{journals/coling/BudanitskyH06} between points by observing their relative closeness, and therefore, topical clusters can be identified.
Moreover, as the directions of highest variance differ from embedding space to embedding space, the projections are incompatible among different embeddings spaces, and this makes them incomparable, a common issue among dimensionality reduction techniques.

t-SNE, differently from PCA, optimizes a loss that encourages embeddings that are in their respective close neighborhoods in the original high-dimensional space to be close in the lower dimensional projection space.
t-SNE projections visually approximate better the original embedding space and topical clusters are more clearly distinguishable, but do not solve the issue of comparability of two different sets of embeddings, nor do they solve the lack of interpretability of the axes or allow for fine-grained inspection.

For these reasons, there is value in mapping embeddings into a more specific, controllable and interpretable semantic space.
In this paper, a new and simple method to inspect, explore and debug embedding spaces at a fine-grained level is proposed.
This technique is made available in Parallax alongside PCA and t-SNE for goal-oriented analysis of the embedding spaces.
It consists of explicitly defining the axes of projection through formulae in vector algebra that use embedding labels as atoms.
Explicit axis definition assigns interpretable and fine-grained semantics to the axes of projection.
This makes it possible to analyze in detail how embeddings relate to each other with respect to interpretable dimensions of variability, as carefully crafted formulas can map (to a certain extent) to semantically meaningful portions of the space.
The explicit axes definition also allows for the comparison of embeddings obtained from different datasets, as long as they have common labels and are equally normalized.

We demonstrate three visualizations that Parallax provides for analyzing subspaces of interest of embedding spaces and a set of example case studies including bias detection, polysemy analysis and fine-grained embedding analysis, but additional ones, like diachronic analysis and the analysis of representations obtained through graph learning or any other means, may be performed as easily.
Moreover, the proposed visualizations can be used for debugging purposes and, in general, for obtaining a better understanding of the embedding spaces learned by different models and representation learning approaches.

The main contribution of this work lies in 1) the tool itself, as Parallax enables researchers in the fields of machine learning, computational linguistics, natural language processing, social sciences and digital humanities to perform exploratory analysis and better understand the semantics of their embeddings. and 2) the use of explicit user-defined algebraic formulae as axes for projecting embedding spaces into semantically-meaningful subspaces that when visualized provide interpretable axes, which allows more fine grained task-oriented analysis and comparison across corpora.

We show how this methodology can be widely used through a series of case studies on well known models and data, and furthermore, we validate its usefulness for goal-oriented analysis through a user study.

\section{Parallax} \label{sec:methodology}

Parallax, is a tool that allows to visualize embedding spaces through the common PCA and t-SNE techniques and through the explicit definition of projection axes through algebraic formulae, which is particularly useful for goal oriented analysis.
Parallax interface, shown in Figure~\ref{fig:parallax_screenshot}, presents a plot on the left side (scatter or polar) and controls on the right side that allow users to define parameters of the projection (what measure to use, values for the hyperparameters, the formuale for the axes in case of explicit axes projections are selected, etc.) and additional filtering and visualization parameters.
Filtering parameters define logic rules applied to embeddings metadata to decide which of them should be visualized, e.g., the user can decide to visualize only the most frequent words or only verbs if metadata about part-of-speech tags is made available.
Filters on the embeddings themselves can also be defined, e.g., the user can decide to visualize only the embeddings with cosine similarity above 0.5 to the embedding of ``horse''.

\begin{figure}
\begin{center}$
\begin{array}{c}
\includegraphics[width=\columnwidth]{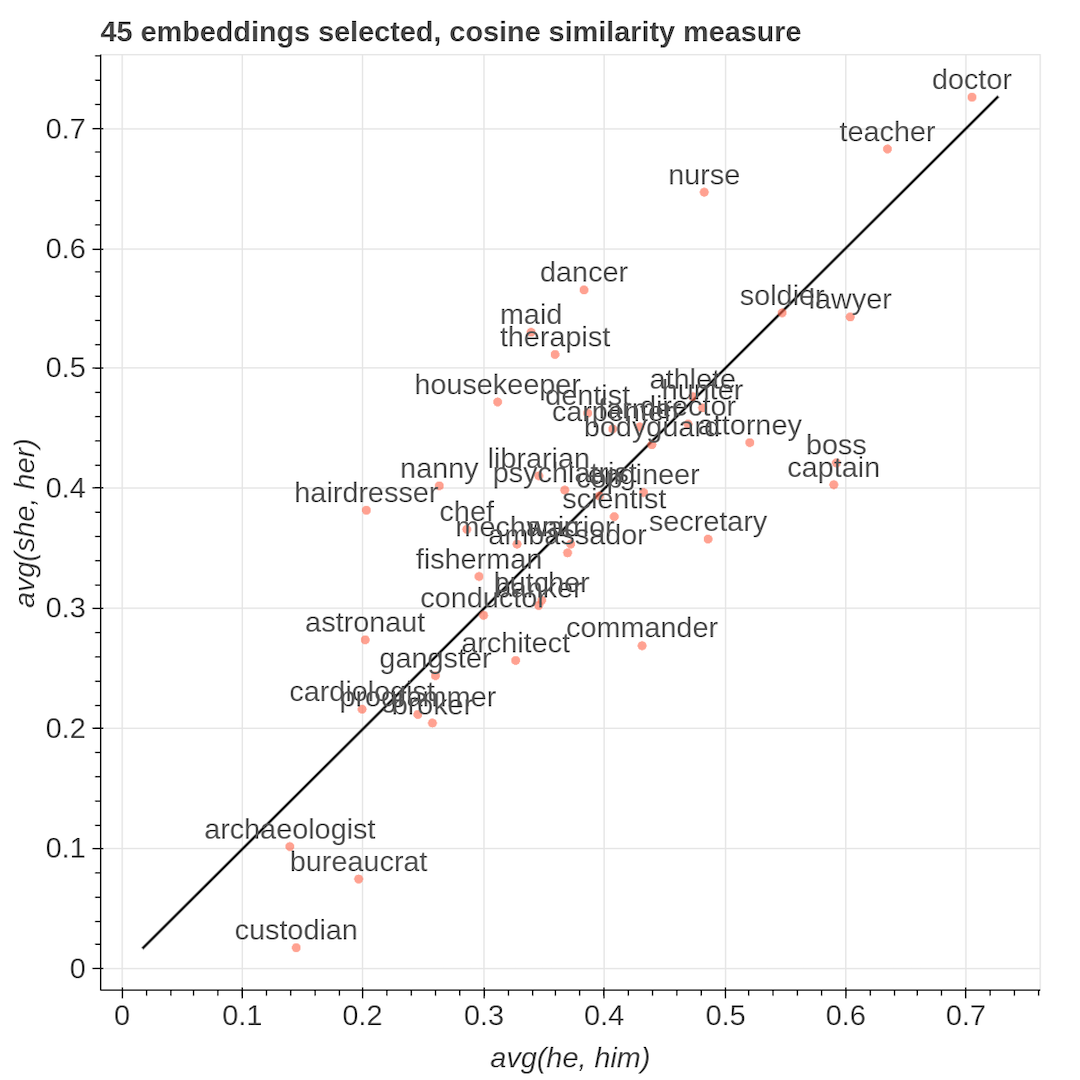} \\
\includegraphics[width=\columnwidth]{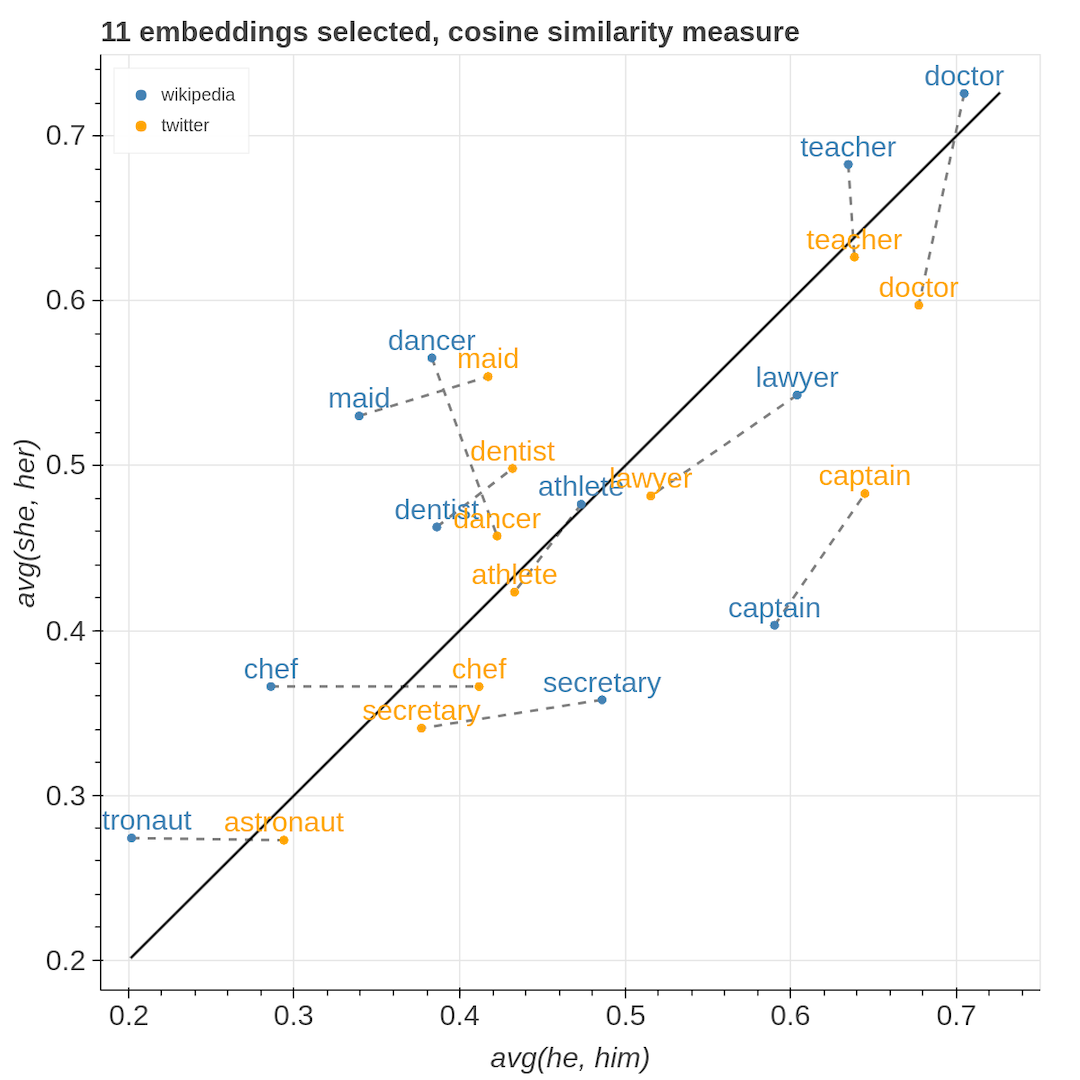}
\end{array}$
\end{center}
\caption{In the top we show professions plotted on ``male'' and ``female'' axes in $Wikipedia$ embeddings. In the bottom we show their comparison in $Wikipedia$ and $Twitter$ datasets.}
\label{fig:professions_cartesian}
\end{figure}

In particular, Parallax's capability of explicitly defining axes is useful for goal-oriented analyses, e.g., when the user has a specific analysis goal in mind, like detecting bias in the embeddings space.
Goals are defined in terms of dimensions of variability (axes of projection) and items to visualize (all the embeddings that are projected, after filtering).
In the case of a few dimensions of variability (up to three) and potentially many items of interest, a Cartesian view is ideal.
Each axis is the vector obtained by evaluating the algebraic formula it is associated with, and the coordinates displayed are similarities or distances of the items with respect to each axis.
Figure~\ref{fig:professions_cartesian} shows an example of a bi-dimensional Cartesian view.
In the case where the goal is defined in terms of many dimensions of variability, a polar view is preferred.
The polar view can visualize many more axes by showing them in a circle, but it is limited in the number of items it can display, as each item will be displayed as a polygon with each vertex lying on a different axis and too many overlapping polygons would make the visualization cluttered.
Figure~\ref{fig:countries_vs_food} shows an example of a five-dimensional polar view.

The use of explicit axes allows for interpretable comparison of different embedding spaces, trained on different corpora or on the same corpora but with different models, or even trained on two different time slices of the same corpora.
The only requirement for embedding spaces to be comparable is that they contain embeddings for all labels present in the formulae defining the axes.
Moreover, embeddings in the two spaces do not need to be of the same dimension, but they need to be normalized.
Items will now have two sets of coordinates, one for each embedding space, and thus they will be displayed as lines.
Short lines are interpreted as items being embedded similarly in the subspaces defined by the axes in both embedding spaces, while long lines are interpreted as really different locations in the subspaces, and their direction gives insight on how items shift in the two subspaces.
Those two embedding spaces could be, for instance, embeddings trained on a clean corpus like Wikipedia as opposed to a noisy corpus like tweets from Twitter, or the two corpora could be two different time slices of the same corpus, in order to compare how words changed over time.
The bottom side of Figure~\ref{fig:professions_cartesian} shows an example of how to use the Cartesian comparison view to compare embeddings in two datasets.

\section{Case Studies} \label{sec:case_studies}

\begin{figure*}
\begin{center}
$
\begin{array}{c}
\includegraphics[width=\textwidth]{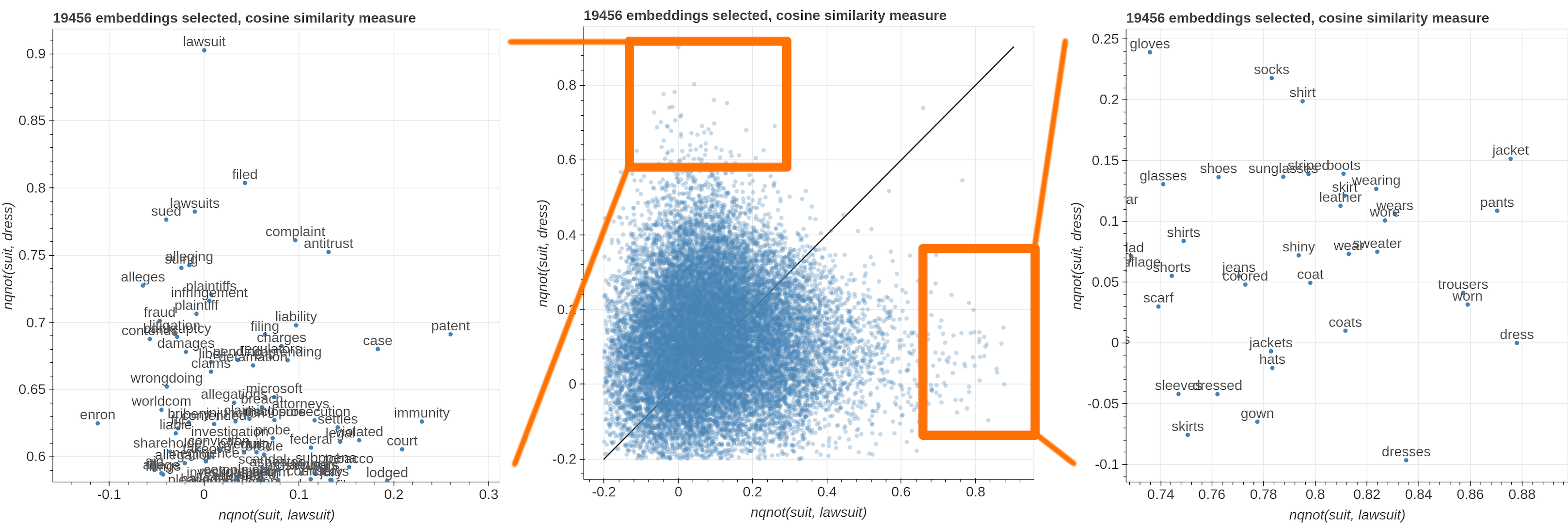} \\
\includegraphics[width=\textwidth]{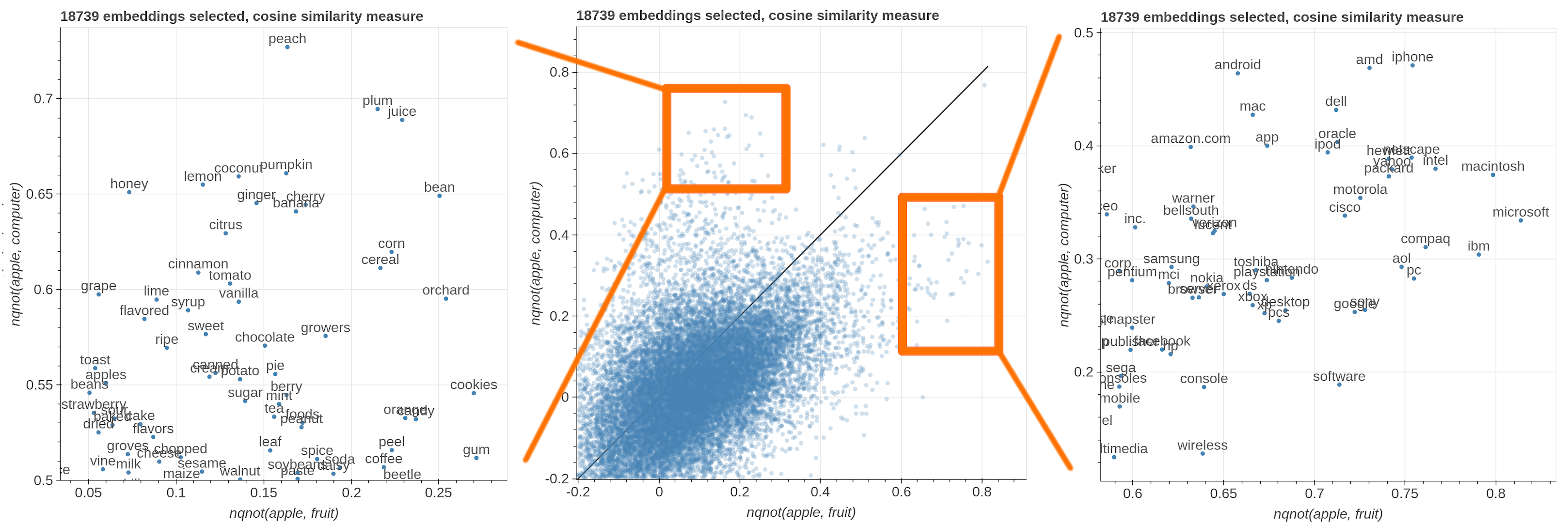}
\end{array}
$
\end{center}
\caption{In the top a plot of embeddings in \textit{Wikipedia} with \textit{suit} negated with respect to \textit{lawsuit} and \textit{dress} respectively as axes. In the bottom a plot of  \textit{apple} negated with respect to \textit{fruit} and \textit{computer}.}
\label{fig:polysemy}
\end{figure*}

\begin{figure}
\begin{center}
$
\begin{array}{c}
\includegraphics[width=\columnwidth]{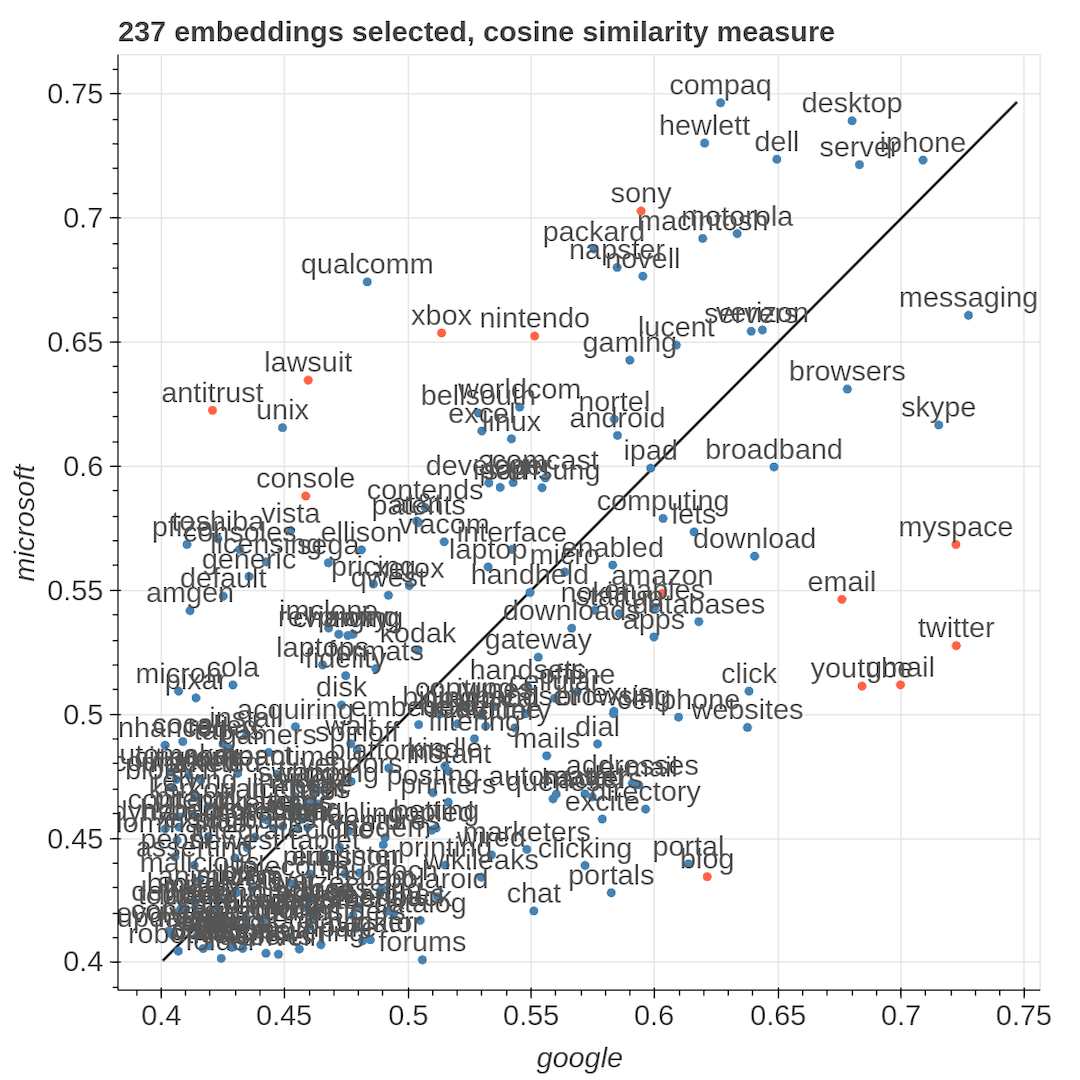} \\
\includegraphics[width=\columnwidth]{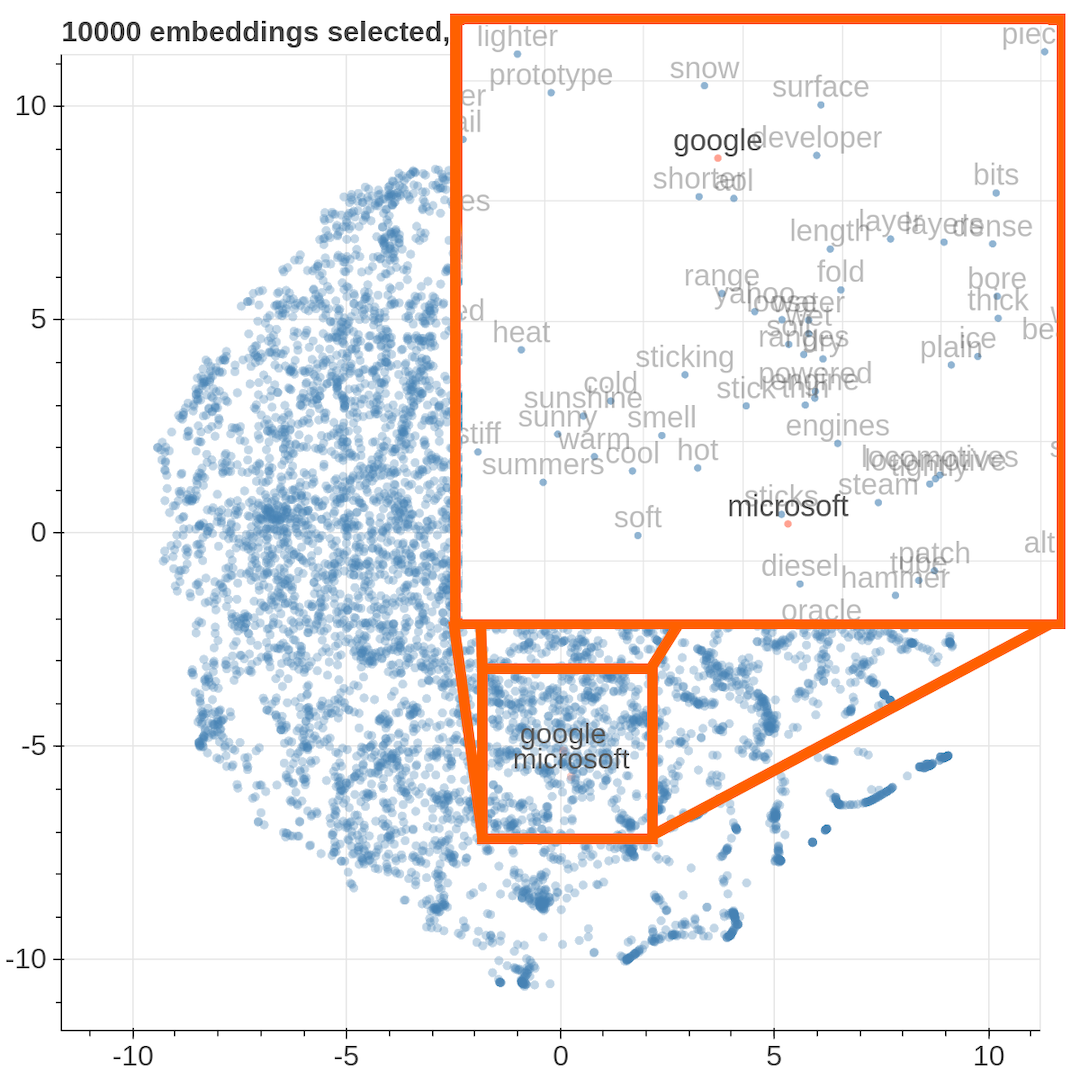}
\end{array}
$
\end{center}
\caption{The top figure is a fine-grained comparison of the subspace on the axis \textit{google} and \textit{microsoft} in \textit{Wikipedia}, the bottom one is the \textit{t-SNE} counterpart.}
\label{fig:google_vs_microsoft}
\end{figure}

\begin{figure*}
\begin{center}
\includegraphics[width=\textwidth]{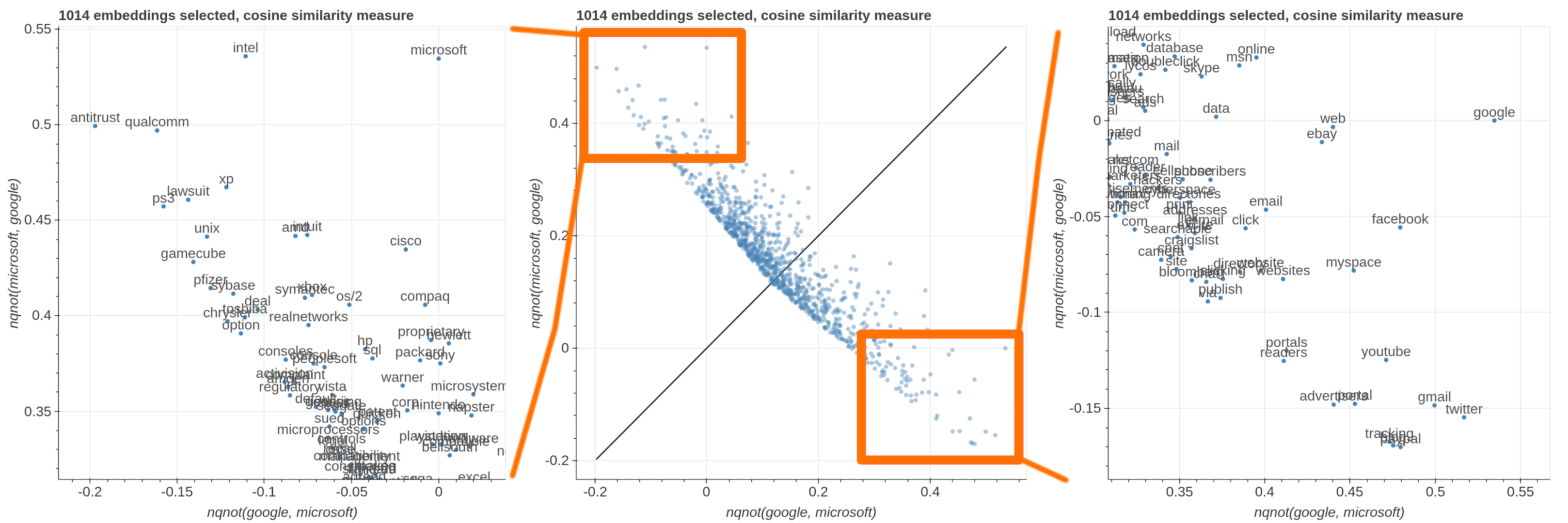}
\end{center}
\caption{Fine-grained comparison of the subspace on the axis $nqnot(google,microsoft)$ and $nqnot(microsoft,google)$ in \textit{Wikipedia}.}
\label{fig:google_vs_microsoft_orth}
\end{figure*}

\begin{figure*}
\begin{center}
\includegraphics[width=\textwidth]{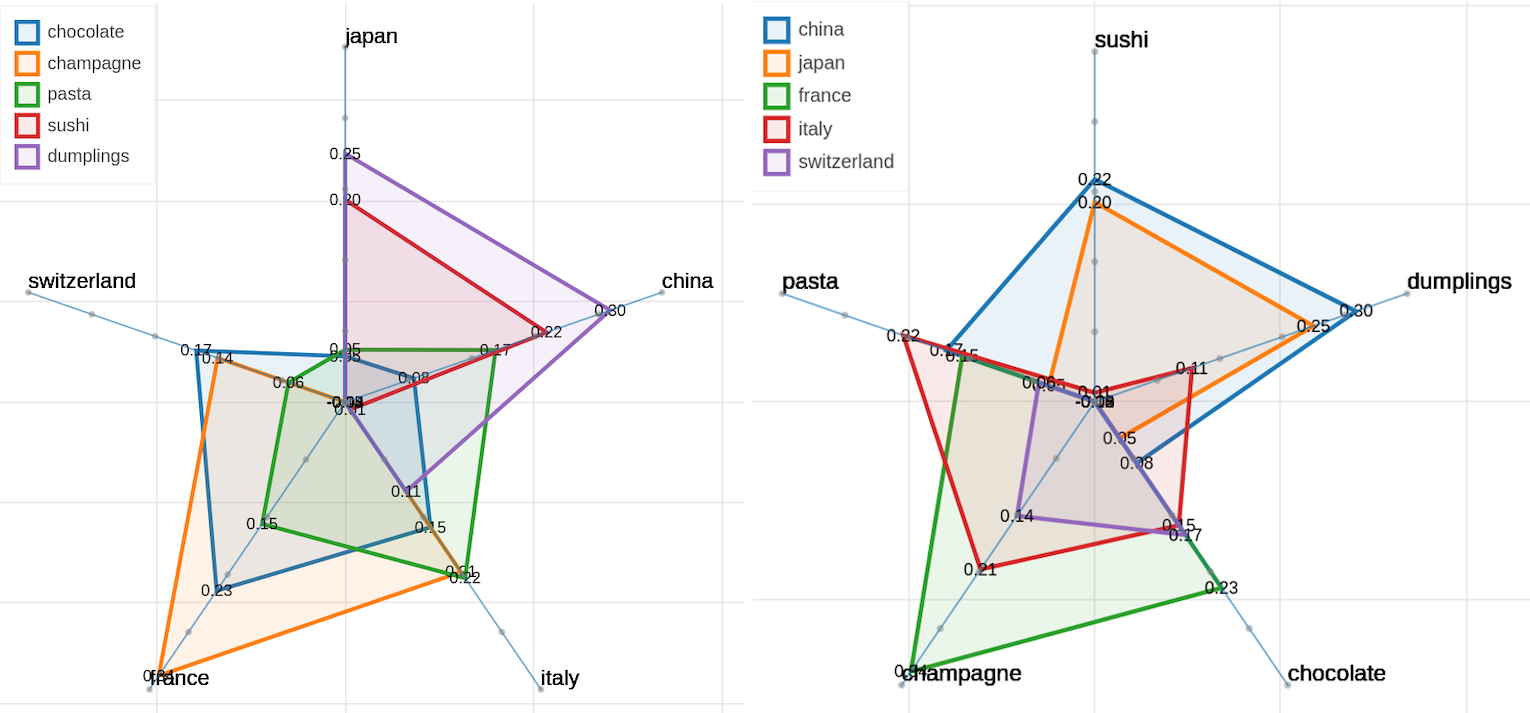}
\end{center}
\caption{Two polar views of countries and foods in \textit{Wikipedia}.}
\label{fig:countries_vs_food}
\end{figure*}

Parallax can be used fruitfully in many analysis tasks in linguistics, digital humanities, in social studies based on empirical methods, and can also be used by researchers in computational linguistics and machine learning to inspect, debug and ultimately better understand the representations learned by their models.

In this section, a few goal-oriented use cases are presented, but Parallax's flexiblity allows for many others.
We used 50-dimensional publicly available GloVe~\citep{pennington2014glove} embeddings trained on Wikipedia and Gigaword 5 summing to 6 billion tokens (for short \textit{Wikipedia}) and 2 billion tweets containing 27 billion tokens (\textit{Twitter}).

\paragraph{Bias detection}

The task of bias detection is to identify, and in some cases correct for, bias in data that is reflected in the embeddings trained on such data.
Studies have shown how embeddings incorporate gender and ethnic biases~(\cite{journals/pnas/GargSJZ18, conf/nips/BolukbasiCZSK16, journals/science/IslamBN16}), while other studies focused on warping spaces in order to de-bias the resulting embeddings~(\cite{conf/nips/BolukbasiCZSK16, conf/emnlp/ZhaoWYOC17}).
We show how our proposed methodology can help visualize biases.

To visualize gender bias with respect to professions, the goal is defined with the formulae $avg(he, him)$ and $avg(she, her)$ as two dimensions of variability, in a similar vein to \cite{journals/pnas/GargSJZ18}.
A subset of the professions used by \cite{conf/nips/BolukbasiCZSK16} is selected as items and cosine similarity is adopted as the measure for the projection.
The Cartesian view visualizing \textit{Wikipedia} embeddings is shown in the left of Figure~\ref{fig:professions_cartesian}.
\textit{Nurse}, \textit{dancer}, and \textit{maid} are the professions closer to the ``female'' axis, while \textit{boss}, \textit{captain}, and \textit{commander} end up closer to the ``male'' axis.

The Cartesian comparison view comparing the embeddings trained on \textit{Wikipedia} and \textit{Twitter} is shown in the right side of Figure~\ref{fig:professions_cartesian}.
Only the embeddings with a line length above $0.05$ are displayed.
The most interesting words in this visualization are the ones that shift the most in the direction of negative slope.
In this case, \textit{chef} and \textit{doctor} are closer to the ``male'' axis in \textit{Twitter} than in \textit{Wikipedia}, while \textit{dancer} and \textit{secretary} are closer to the bisector in \textit{Twitter} than in \textit{Wikipedia}.

Additional analysis of how words tend to shift in the two embedding spaces would be needed in order to derive provable conclusions about the significance of the shift, for instance through a permutation test with respect to all possible pairs, but the visualization can help inform the most promising words to perform the test on.

\paragraph{Polysemy analysis} \label{sec:polysemy_analysis}

Methods for representing words with multiple vectors by clustering contexts have been proposed~\citep{conf/acl/HuangSMN12, conf/emnlp/NeelakantanSPM14}, but widely used pre-trained vectors conflate meanings in the same embedding.

\citet{conf/acl/Widdows03} showed how using a binary orthonormalization operator that has ties with the quantum logic \textit{not} operator it is possible to remove part of the conflated meaning from the embedding of a polysemous word.
The authors define the operator $nqnot(a,b) = a - \frac{a \cdot b}{|b|^2}b$ and we show with a comparison plot how it can help distinguish the different meanings of a word.

For illustrative purposes, we choose the same polysemous word used by \cite{conf/acl/Widdows03}, \textit{suit}, and use the $nqnot$ operator to orthonormalize with respect to \textit{lawsuit} and \textit{dress}, the two main meanings used as dimensions of variability.
The items in our goal are the 20,000 most frequent words in the \textit{Wikipedia} embedding space removing stop-words.
In Figure~\ref{fig:polysemy}, we show the overall plot and we zoom on the items that are closer to each axis.
Words closer to the axis negating \textit{lawsuit} are all related to dresses and the act of wearing something, while words closer to the axis negating \textit{dress} are related to law.

We chose another polysemous word, \textit{apple}, and orthonornalized with respect to \textit{fruit} and \textit{computer}.
In the bottom of Figure~\ref{fig:polysemy} words that have a higher similarity with respect to the first axis are all tech related, while the ones that have a higher similarity with respect to the second axis are mostly other fruits or food.

Both examples confirm the ability of the $nqnot$ operator to disentangle multiple meanings from polysemous embeddings and show how the proposed visualizations are able to show it clearly.

\paragraph{Fine-grained embedding analysis} \label{sec:finegrained_analysis}

We consider embeddings that are close to be semantically related, but even close embeddings may have nuances that distinguish them.
When projecting in two dimensions through PCA or t-SNE we are conflating a multidimensional notion of similarity to a bi-dimensional one, losing the fine-grained distinctions.
The Cartesian view allows for a more fine-grained visualization that emphasizes nuances that could otherwise go unnoticed.

To demonstrate this capability, we select as dimensions of variability single words in close vicinity to each other in the \textit{Wikipedia} embedding space: \textit{google} and \textit{microsoft}, as \textit{google} is the closest word to \textit{microsoft} and \textit{microsoft} is the 3\textsuperscript{rd} closest word to \textit{google}.
As items, we pick the 30,000 most frequent words removing stop-words and remove the 500 most frequent words (as they are too generic) and keeping only the words that have a cosine similarity of at least $0.4$ with both \textit{google} and \textit{microsoft} and a cosine similarity below $0.75$ with respect to $google+microsoft$, as we are interested in the most polarized words.

The left side of Figure~\ref{fig:google_vs_microsoft} shows how even if those embeddings are close to each other, it is easy to identify peculiar words (highlighted with red dots).
The ones that relate to web companies and services (\textit{twitter}, \textit{youtube}, \textit{myspace}) are much closer to the \textit{google} axis.
Words related to both legal issues (\textit{lawsuit}, \textit{antitrust}) and videogames (\textit{ps3}, \textit{nintendo}, \textit{xbox}) and traditional IT companies are closer to the \textit{microsoft} axis.

In Figure~\ref{fig:google_vs_microsoft_orth} we obtain similar results by using \textit{google} and \textit{microsoft} orthonormalized with respect to each other as axes.
The top left and the bottom right corners are the most interesting ones, as they contain terms that are related to one word after having negated the other.
The pattern that emerges is similar to the one highlighted in the left side of Figure~\ref{fig:google_vs_microsoft}, but now also operating systems terms (\textit{unix}, \textit{os/2}) appear in the \textit{microsoft} corner, while \textit{advertisement} and \textit{tracking} appear in the \textit{google} corner.

For contrast, the t-SNE projection is shown in the right side of Figure~\ref{fig:google_vs_microsoft}: it is hard to appreciate the similarities and differences among those embeddings other than seeing them being close in the projected space.
This confirms on one hand that the notion of similarity between terms in an embedding space hides many nuances that are captured in those representations, and on the other hand, that the proposed methodology enables for a more detailed inspection of the embedded space.

Multi-dimensional similarity nuances can be visualized using the polar view.
In Figure~\ref{fig:countries_vs_food}, we show how to use Parallax to visualize a small number of items on more than two axes, specifically five food-related items compared over five countries' axes.
The most typical food from a specific country is the closest to the country axis, with \textit{sushi} being predominantly close to \textit{Japan} and \textit{China}, \textit{dumplings} being close to both Asian countries and \textit{Italy}, \textit{pasta} being predominantly closer to \textit{Italy}, \textit{chocolate} being close to European countries and \textit{champagne} being closer to \textit{France} and \textit{Italy}.
This same approach could be also be used for bias detection among different ethnicities, for instance, where the axes are concepts capturing the notion of ethnicity and items could be adjectives, or the two could be swapped, depending.

\paragraph{Analogy} \label{sec:analogy}

\begin{figure}

\includegraphics[width=\columnwidth]{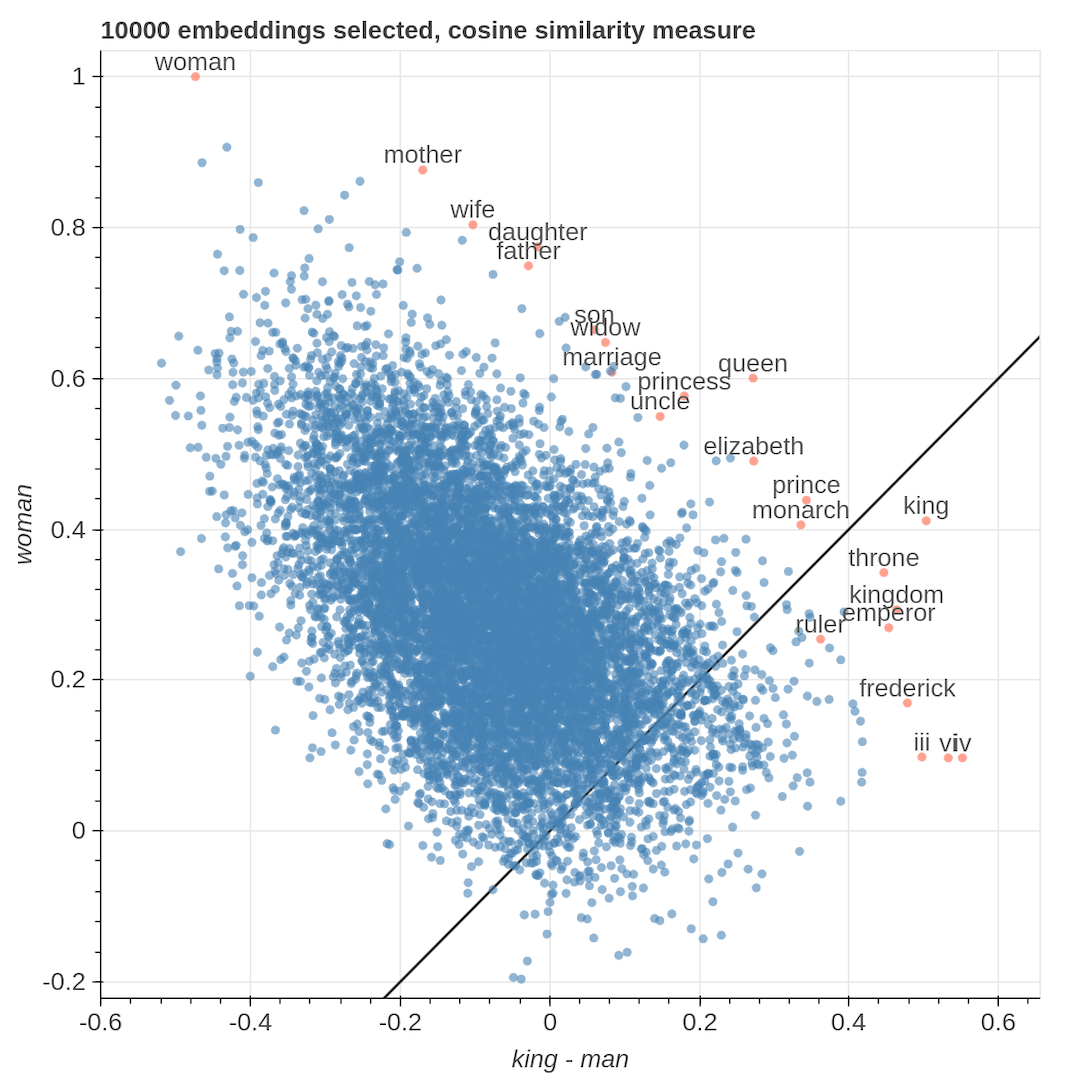}
\caption{king-man vs woman}
\label{fig:king-man_vs_woman}
\end{figure}

\begin{figure}
\includegraphics[width=\columnwidth]{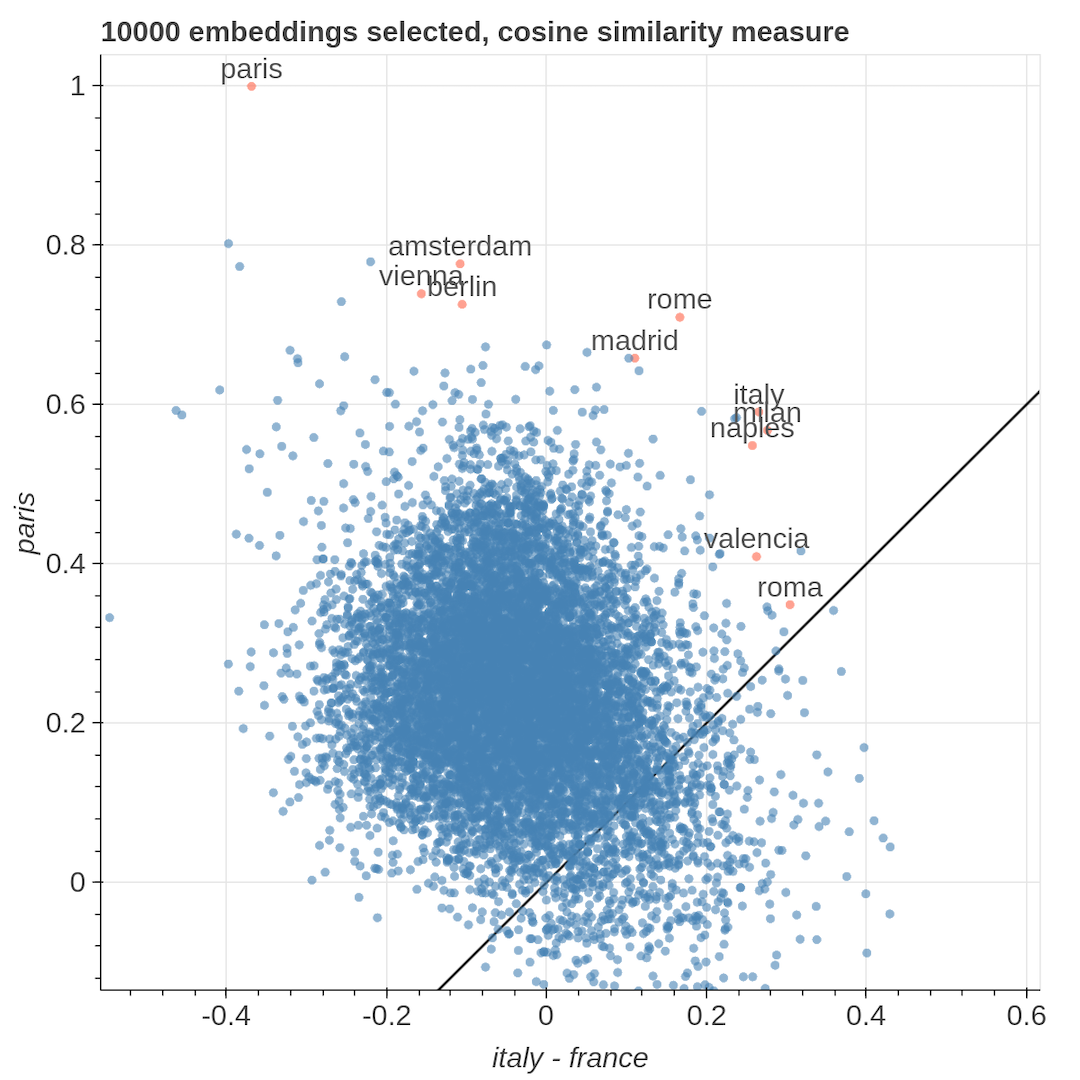}
\caption{italy-france vs paris}
\label{fig:italy-france_vs_paris}
\end{figure}

The analogy task introduced by \cite{conf/nips/MikolovSCCD13} is often used as an example to show the amount of information encoded in the dimensions of the embedding space and the fact that linear relationships emerge from training embedding models.
In this section we show how using explicit axes makes it really easy to analyze more in detail the space of linear relationships.

Two examples are presented.

In Figure~\ref{fig:king-man_vs_woman} we show a cartesian plot where the axes are 'king-man' and 'woman' respectively.
The bisector line depicted is the direction of the sum, meaning that word that are far off in the direction (closer to both axes) are to be considered good candidates to solve the analogy.
In this case the word that are further in the direction are 'king' and 'queen', but usually when computing scores in these tasks, words already present in the analogy are omitted, so, removing 'king', 'queen' remain as the best candidate for the analogy.
In the plot we highlight several words in the same band perpendicular to the bisector, meaning that all those words are similarly good candidates for the analogy.
Within the band, some words are closer to the `king-man` axis (interpretable as the concept of 'royalty'), for instance: 'frederick', 'emperor', 'ruler', throne', but also roman numerals like 'iii', 'iv' and 'vi'.
Other words within the band are close to the 'queen' axis: 'mother', 'wife', 'daughter' and 'father'.

In Figure~\ref{fig:italy-france_vs_paris} we show a cartesian plot where the axes are 'italy-france' and 'paris' respectively.
The same interpretation of the bisector and the bands of the previous example hold also in this case.
'rome' is the word further off in both directions, while words closer to the `paris` axis tend to be other European capitals: 'amsterdam, 'vienna', 'berlin'.
Words closer to the `italy-rome` axis (interpretable as ``italianness'') are mostly Italian cities: 'italy', 'naples', 'milan', 'valencia'.

This way of visualizing analogies provides a richer understanding of the embedding space and a characterization of why some words appear in the analogy ranking list.

\paragraph{Paradigmatic versus Syntagmatic} \label{sec:paradisgmatic_vs_syntagmatic}

\begin{figure}
\includegraphics[width=\columnwidth]{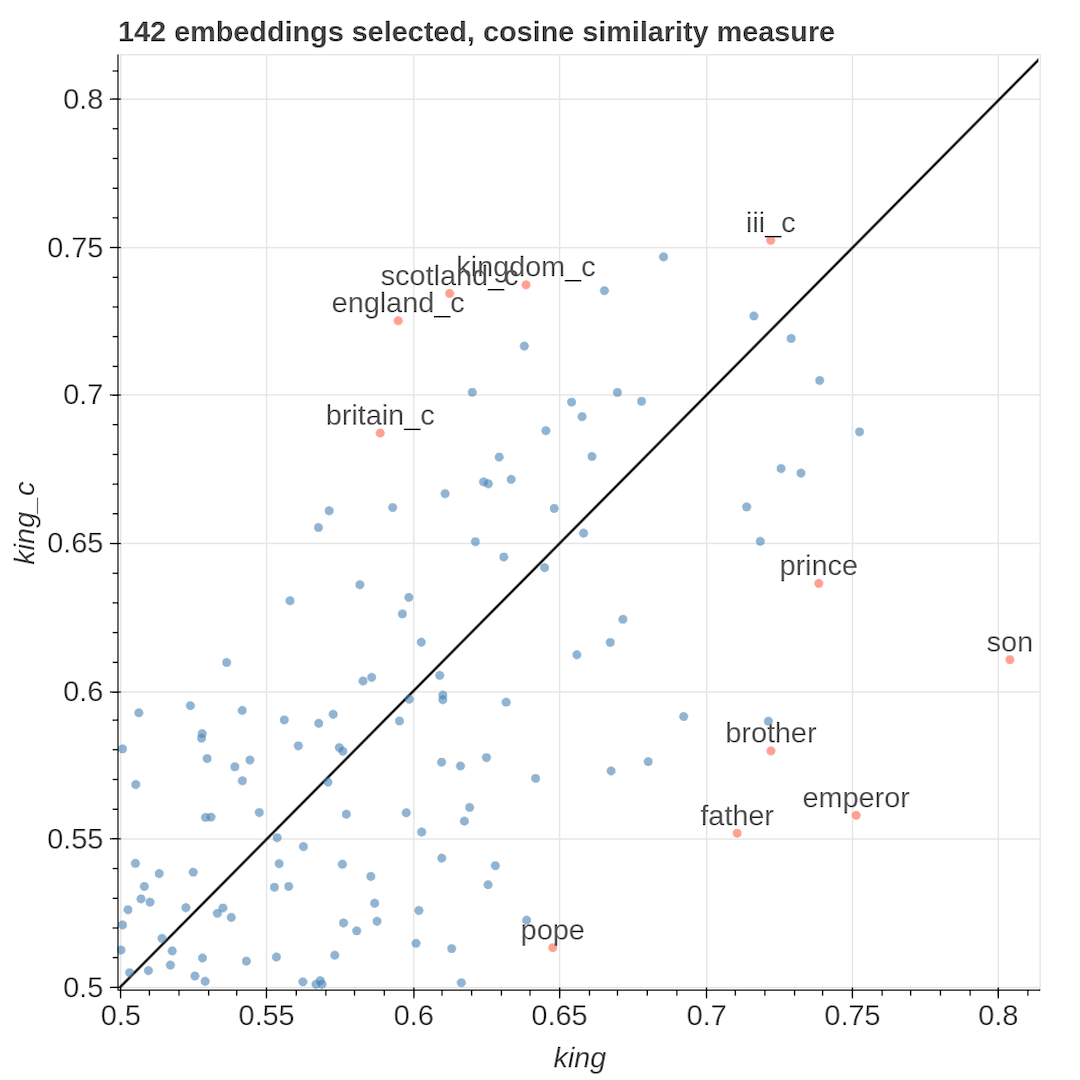}
\caption{paradisgmatic vs syntagmatic}
\label{fig:paradisgmatic_vs_syntagmatic}
\end{figure}

Depending on the definition of context embeddings methods can learn to a different notion of relatedness in the similarity of the vectors that are obtained.
Even with a specific notion of context, such as a window of occurrence, the model architecture can lead to obtain representations that encode different word relationships.
For instance, the GloVe \citep{pennington2014glove} algorithm trains two sets of vectors, one for the rows of the co-occurrence matrix and one for the columns.
Those two sets of vectors will likely contain different information as embeddings of row words will be similar for words that co-occur with the same contexts and are then substitutable with each other (paradigmatic relationship), while embeddings of the column words will be similar to each other if they appear co-occur together in the contexts (syntagmatic relationship).
At the end of the training of a GloVe model, the two sets of vectors are summed together, obfuscating their different nature.
We modified the GloVe training procedure to save the two embeddings set without summing them and trained a model on the text8 dataset as a proof of concept.

In Figure~\ref{fig:king-paradisgmatic_vs_syntagmatic} we show a cartesian plot using the word `king` (the embedding of the row word) and `king\_c` (the embedding of the row word) as axes.
Note that `\_c` at the end of a word denotes an embedding of a column, it is just a simple way to differentiate among the two sets of vectors.
By looking at the word that are closer to the `king` axis, `emperor`, `prince`, `pope`, `father` and `son` it is evident that they are indeed words in a paradigmatic relationship with `king` as they are substitutable.
On the other hand, looking at the words closer to the `king\_c` axis, the word `scotland\_c`, `kingdom\_c`, `england\_c`, `britain\_c` and `iii\_c` appear.
Those words are cleary in a syntagmatic relationship with `king\_c` as the will probably appear in the same window of text like `the King of England` or `King George III`.

With the use of explicit axes this phenomenon is made clear and visible.

\paragraph{Knowledge Base Embeddings} \label{sec:knowledge_base_embeddings}

\begin{figure}
\includegraphics[width=\columnwidth]{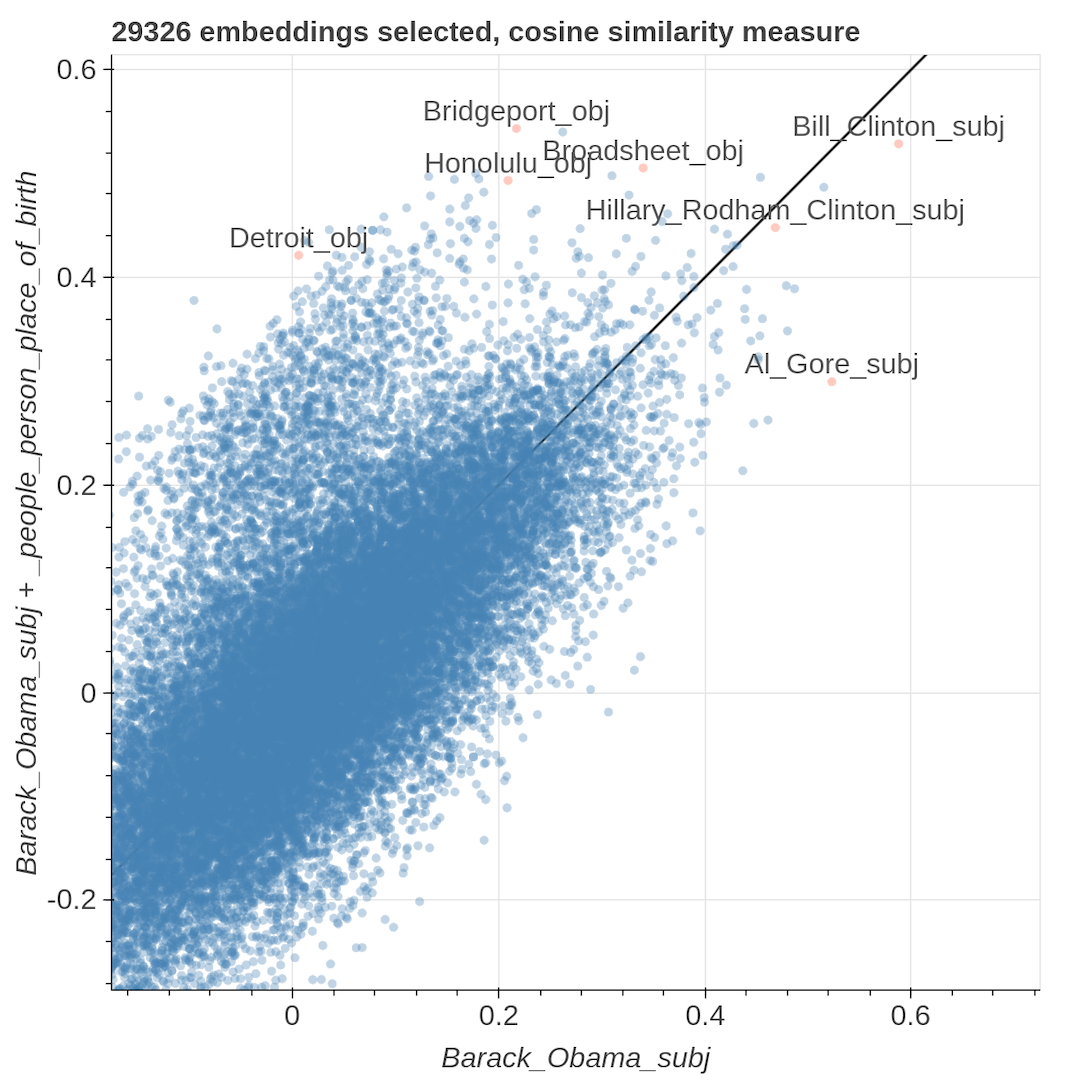}
\caption{knowledge bases}
\label{fig:knowledge_bases}
\end{figure}

Another area where embeddings are trained is automatic construction of Knowledge Bases (KBs).
In this thread of works, representations of entities in the KB and their relationships are encoded, with the aim to use those representations to automatically infer missing data and in some cases populate a probabilistic KB.

Many approaches have been proposed, the one we use here in order to obtain some insights through our proposed visualizations is StarSpace~\cite{WuFCABW18}, trained on the FB15K dataset \citep{BordesUGWY13}, a subset of FreeBase.
In this algorithm different embeddings are obtained for each subject, predicate and object by training to predict the object from the subject and the predicate.
Given the triple <Barack\_Obama, place\_of\_birth, Honolulu>, the model will minimize the distance between $e_{subj}(Barack\_Obama) + e_{pred}(place\_of\_birth)$ and $e_{obj}(Honolulu)$, where $e_{subj}$, $e_{pred}$ and $e_{obj}$ are functions that map identifiers of entities and predicates to their embeddings when they appear as subject, predicate and object respectively.
This also implies that the same entity may have distinct subject and object embeddings.

In Figure~\ref{fig:knowledge_bases} we show a cartesian plot using the entity `Barack\_Obama\_subj` (the embedding of the entitiy Barack\_Obama when it appears as subject) and `Barack\_Obama\_subj + person\_person\_place\_of\_birth` (the of the embedding of the entity Barack\_Obama and the predicate person\_person\_place\_of\_birth that denotes the relationship between people and their place of birth) as axes.
Ideally the second axis should be close to the place of birth of Barack Obama, as that is what the model was optimized for.
From the visualization it is apparent how the embeddings close to the Barack\_Obama\_subj axis below or close to the bisector are mostly other American democratic presidents or candidate presidents, while the ones with higher cosine similarity with respect to the vertical axis are mostly cities.
The correct object, Honolulu, is among the top ranked ones, although not the one with the highest similarity to the embedding of the vertical axis.



\newcommand{\Proj}{Projection\xspace}
\newcommand{\ProjE}{Explicit Formulae\xspace}
\newcommand{\ProjT}{t-SNE\xspace}
\newcommand{\Task}{Task\xspace}
\newcommand{\TaskC}{Commonality\xspace}
\newcommand{\TaskP}{Polarization\xspace}
\newcommand{\Encode}{Obfuscation\xspace}
\newcommand{\EncodeObf}{Obfuscated\xspace}
\newcommand{\EncodeRaw}{Non-obfuscated\xspace}

\begin{table}
\centering
\small{
  \begin{tabular*}{\columnwidth}{@{}r|lcl@{}}
  \toprule
  Accuracy     & Factor               & $F_{(1,91)}$ & p-value       \\
  \midrule   
  \Proj        & \Proj                & $46.11$      & $0.000^{***}$ \\
  $\times$     & \Task                & $1.709$      & $0.194$       \\
  \Task        & \Proj $\times$ \Task & $3.452$      & $0.066$       \\
  
  \midrule   
  \Proj       & \Proj                  & $57.73$      & $0.000^{***}$ \\
  $\times$    & \Encode                & $23.93$      & $0.000^{***}$ \\
  \Encode     & \Proj $\times$ Obf & $5.731$      & $0.019^{*}$   \\
  \bottomrule
  \end{tabular*}
\caption{Two-way ANOVA analyses of \Task (\TaskC vs. \TaskP) and \Encode (\EncodeObf vs. \EncodeRaw) over \Proj (\ProjE vs. \ProjT).}
\label{table:two_way_anova}
}
\end{table}

\section{User Study}

We conducted a user study to find out if and how visualizations using user-defined semantically meaningful algebraic formulae help users achieve their analysis goals.
What we are not testing for is the projection quality itself, as in PCA and t-SNE it is obtained algorithmically, while in our case it is explicitly defined by the user.
We formalized the research questions as: 
Q1) Does \ProjE outperform \ProjT in goal-oriented tasks? 
Q2) Which visualization do users prefer?

To answer these questions we invited twelve subjects among data scientists and machine learning researchers, all acquainted with interpreting dimensionality reduction results.
We defined two types of tasks, namely \TaskC and \TaskP, in which subjects were given a visualization together with a pair of words (used as axes in \ProjE or highlighted with a big font and red dot in case of \ProjT).
We asked the subjects to identify either common or polarized words w.r.t. the two provided ones.
The provided pairs were: banana \& strawberry, google \& microsoft, nerd \& geek, book \& magazine.
The test subjects were given a list of eight questions, four per task type, and their proposed lists of five words are compared with a gold standard provided by a committee of two computational linguistics experts.
The tasks are fully randomized within the subject to prevent from learning effects.
In addition, we obfuscated half of our questions by replacing the words with a random numeric ID to prevent prior knowledge from affecting the judgment.
We track the \textit{accuracy} of the subjects by calculating the number of words provided that are present in the gold standard set, and we also collected an overall \textit{preference} for either visualizations.

As reported in Table~\ref{table:two_way_anova}, two-way ANOVA tests revealed significant differences in accuracy for the factor of \Proj and \ProjT against both \Task and \Encode, which is a strong indicator that the proposed \ProjE method outperforms \ProjT in terms of accuracy in both \TaskC and \TaskP tasks.
We also observed significant differences in \Encode: subjects tend to have better accuracy when the words are not obfuscated.
We run post-hoc t-tests that confirmed how the accuracy of \ProjE on \EncodeRaw is significantly better than \EncodeObf, which in turn is significantly better that \ProjT \EncodeRaw, which is significantly better than \ProjT \EncodeObf.
Concerning Preference, nine out of all twelve (75\%) subjects chose \ProjE over \ProjT.
In conclusion, our answers to the research questions are that 
(Q1) \ProjE leads to better accuracy in goal-oriented tasks, 
(Q2) users prefer \ProjE over \ProjT.

\section{Related Work}

\textbf{Embedding methods and applications}.
Several methods for learning embeddings from symbolic data have been recently proposed~\citep{pennington2014glove, conf/nips/MikolovSCCD13, conf/nips/MnihK13, Lebret2014WordET, conf/emnlp/JiYYMV16, conf/nips/RudolphRMB16, journals/pieee/Nickel0TG16}.
The learned representations have been used for a variety of tasks like recommendation~\citep{conf/mlsp/BarkanK16}, link prediction on graphs~\citep{conf/kdd/GroverL16}, discovery of drug-drug interaction~\citep{journals/ws/AbdelazizFHZS17} and many more.
In particular, positive results in learning embeddings for words using a surrogate prediction task~\citep{conf/nips/MikolovSCCD13} started the resurgence of interest in those methods, while a substantial body of research from the distributional semantics community using count and matrix factorization based methods~\citep{journals/jasis/DeerwesterDLFH90, journals/coling/BaroniL10, Kanerva00randomindexing, conf/conll/LevyG14, journals/jlm/BiemannR13} was previously developed. Refer to \cite{journals/annrevling/Lenci2018} for a comprehensive overview.
In some of those papers, explicit axes are used to visualize portions of the embedding space in an ad-hoc fashion.

In their recent paper, \cite{HG18} extracted a list of routinely conducted tasks where embeddings are employed in visual analytics for NLP, such as \textit{compare concepts}, \textit{finding analogies}, and \textit{predict contexts}.
iVisClustering~\citep{Lee2012} represents topic clusters as their most representative keywords and displays them as a 2D scatter plot and a set of linked visualization components supporting interactively constructing topic hierarchies.
ConceptVector~\citep{Park2018} makes use of multiple keyword sets to encode the relevance scores of documents and topics: positive words, negative words, and irrelevant words.
It allows users to select and build a concept iteratively.
\cite{Liu2018} display pairs of analogous words obtained through analogy by projecting them on a 2D plane obtained through a PCA and an SVM to find the plane that separates words on the two sides of the analogy.
Besides word embeddings, visualization has been used to understand topic modeling~\citep{Chuang2012} and how topic models evolve over time~\citep{Havre2002}.
Compared to existing literature, our work allows for more fine-grained direct control over the conceptual axes and the filtering logic, allowing users to: 1) define concepts based on explicit algebraic formulae beyond single keywords, 2) filter depending on metadata, 3) perform multidimensional projections beyond the common 2D scatter plot view using the polar view, and 4) perform comparisons between embeddings from different data sources.
Those features are absent in other proposed tools.

\section{Conclusions}

In this work, we presented Parallax, a tool for embedding visualization, and a simple methodology for projecting embeddings into lower-dimensional semantically-meaningful subspaces through explicit algebraic formulae.
We showed how this approach allows goal-oriented analyses and more fine-grained and cross-dataset comparisons through a series of case studies and a user study.

\section*{Acknowledgments}
The authors want to thank Antonio Vergari, Eli Bingham, Fritz Obermayer, Gaetano Rossiello, Pasquale Minervini, Lezhi Li, Zoubin Gahrhamani and Peter Dayan for the fruitful conversations and opinions that lead to improvement of this work.

\bibliography{bibliography}
\bibliographystyle{acl_natbib}

\appendix
\section{Appendix}

\begin{figure*}
\begin{center}
\includegraphics[width=\textwidth]{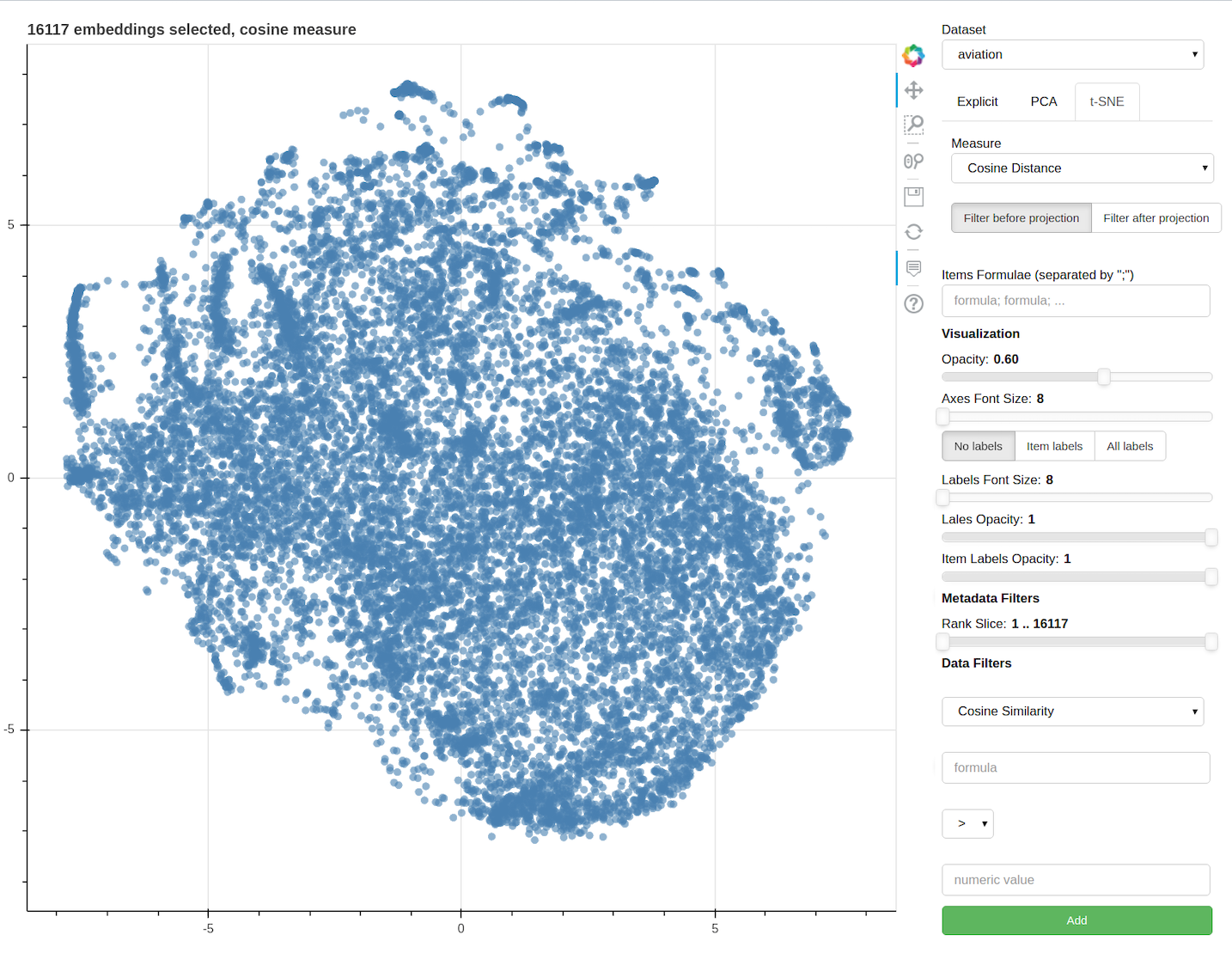}
\end{center}
\caption{Screenshot of Parallax.}
\end{figure*}

\clearpage

\begin{figure*}
\includegraphics[width=\textwidth]{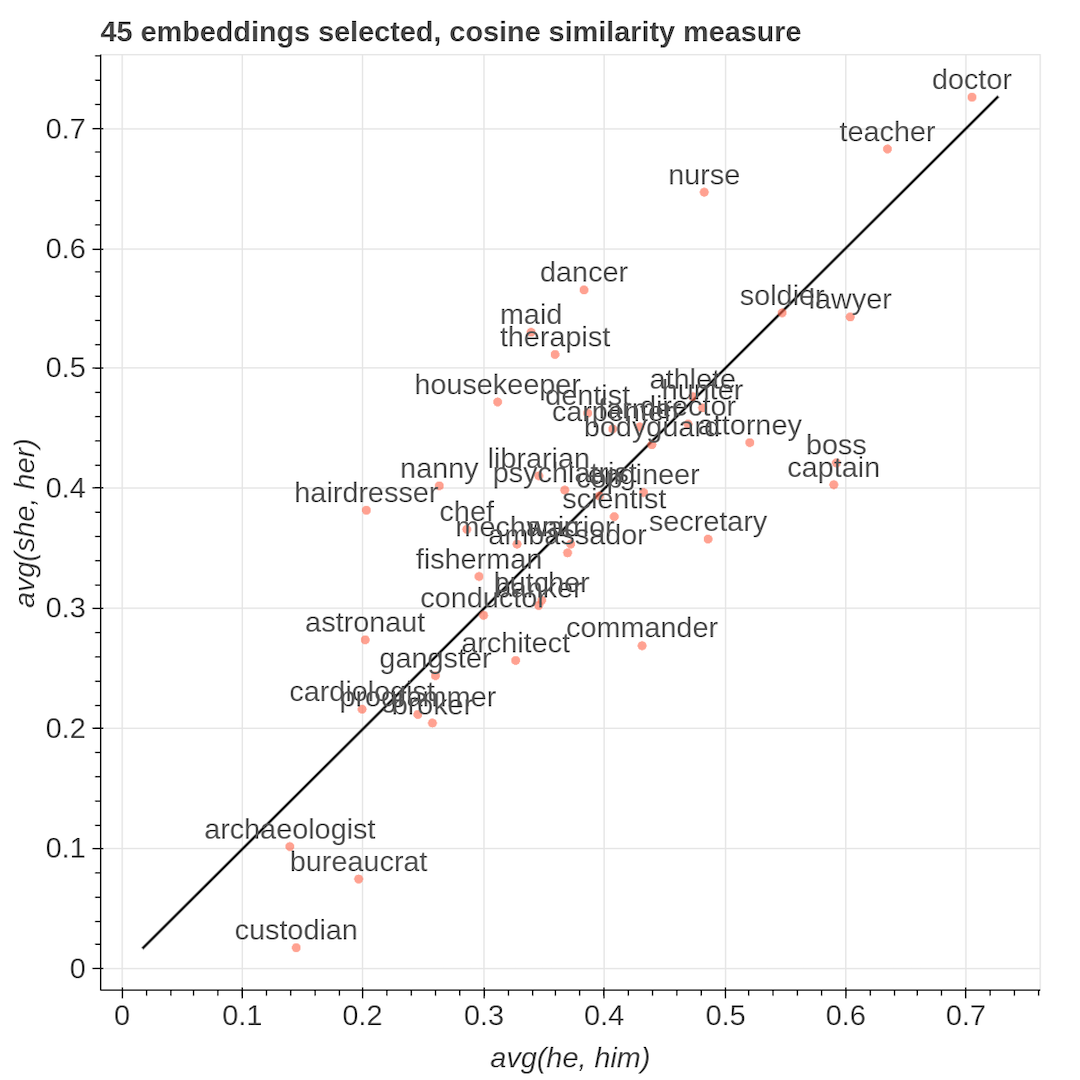}
\caption{Professions plotted on ``male'' and ``female'' axes in $Wikipedia$ embeddings.}
\end{figure*}

\clearpage

\begin{figure*}
\includegraphics[width=\textwidth]{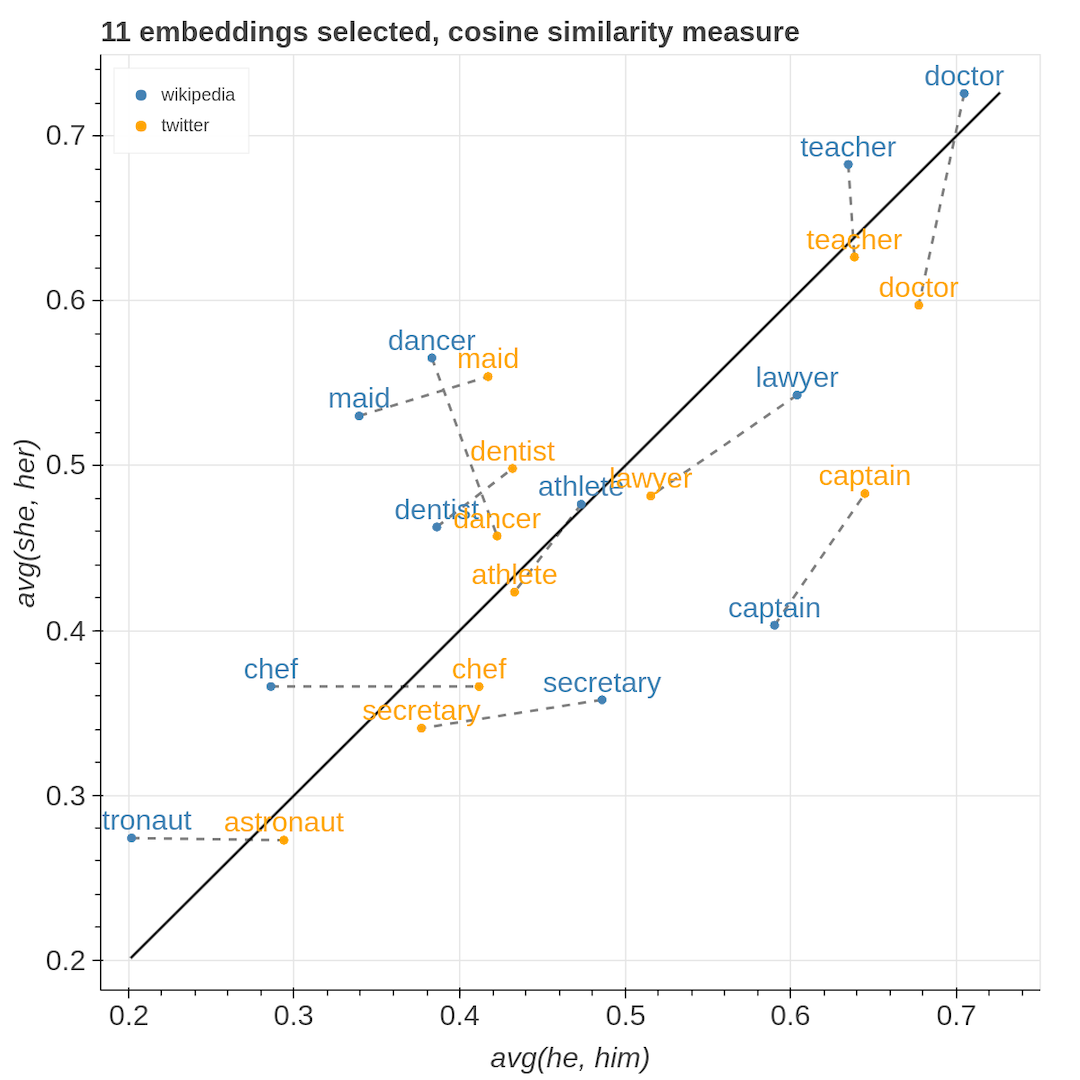}
\caption{Professions plotted on ``male'' and ``female'' axes in $Wikipedia$ and $Twitter$ embeddings.}
\end{figure*}

\clearpage

\begin{sidewaysfigure*}
\includegraphics[width=\textheight]{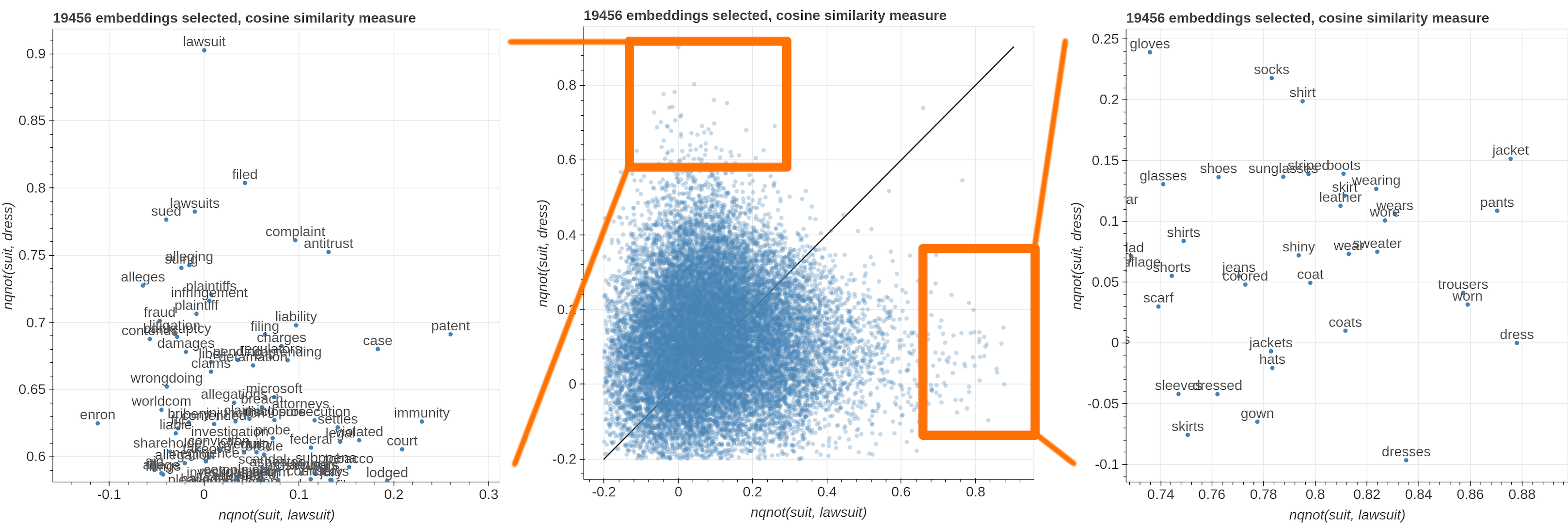}
\caption{Plot of embeddings in \textit{Wikipedia} with \textit{suit} negated with respect to \textit{lawsuit} and \textit{dress} respectively as axes.}
\end{sidewaysfigure*}

\clearpage

\begin{sidewaysfigure*}
\includegraphics[width=\textheight]{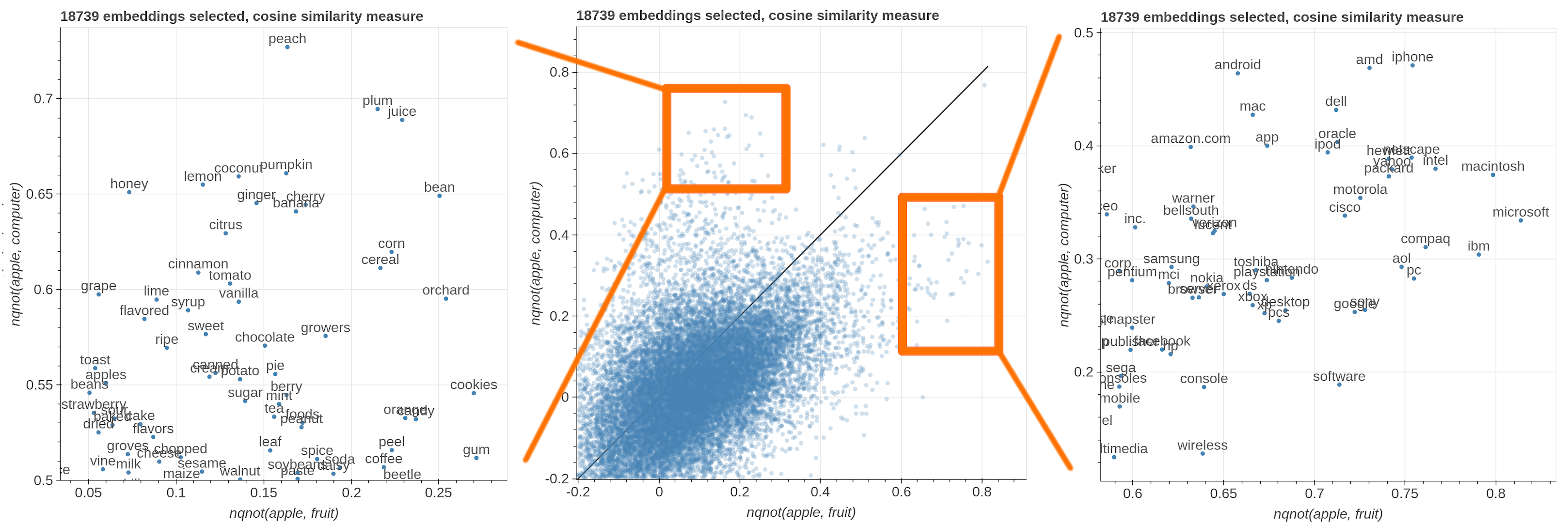}
\caption{Plot of embeddings in \textit{Wikipedia} with \textit{apple} negated with respect to \textit{fruit} and \textit{computer} respectively as axes.}
\end{sidewaysfigure*}

\clearpage

\begin{figure*}
\includegraphics[width=\textwidth]{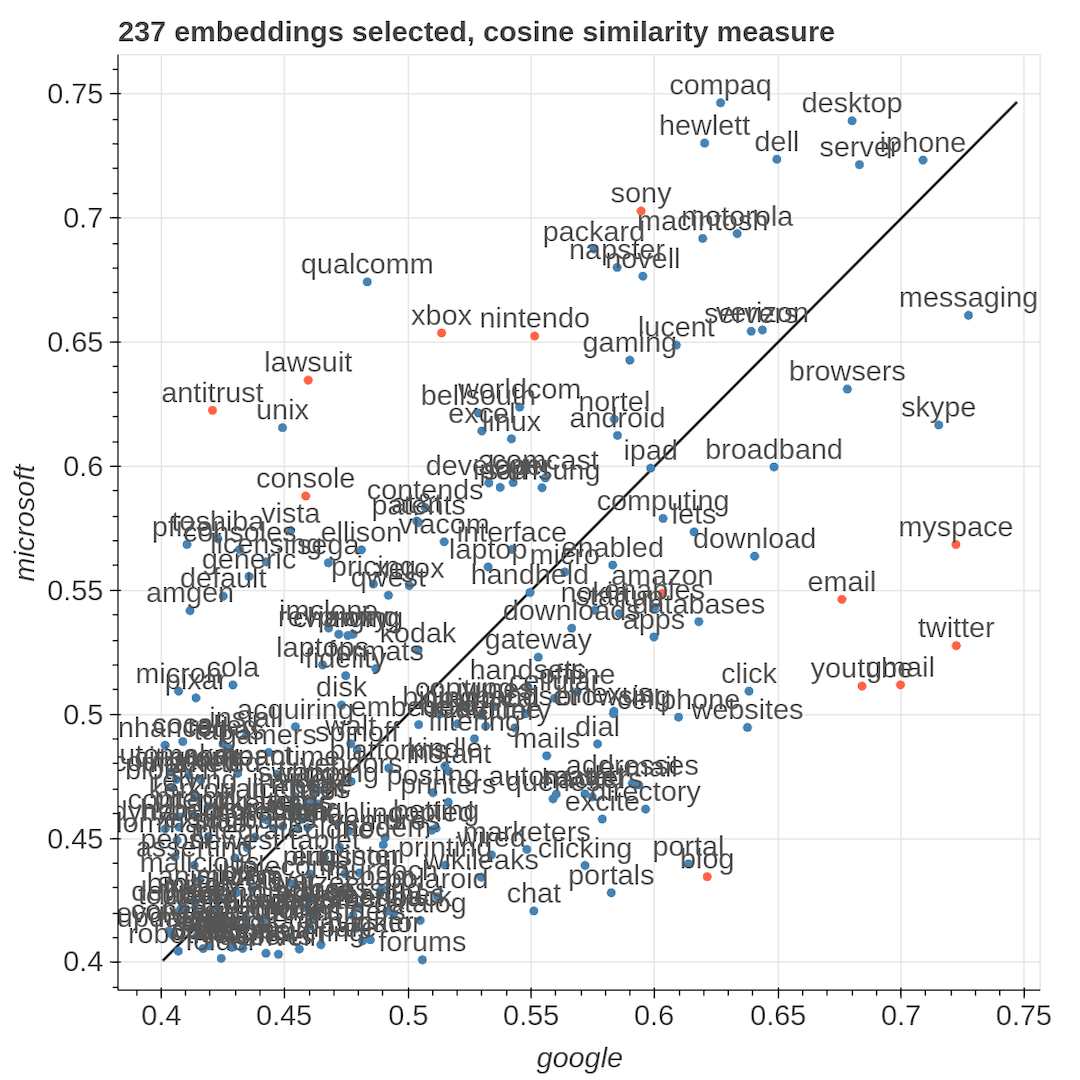}
\caption{Fine-grained comparison of the subspace on the axis \textit{google} and \textit{microsoft} in \textit{Wikipedia}.}
\end{figure*}

\clearpage

\begin{sidewaysfigure*}
\includegraphics[width=\textheight]{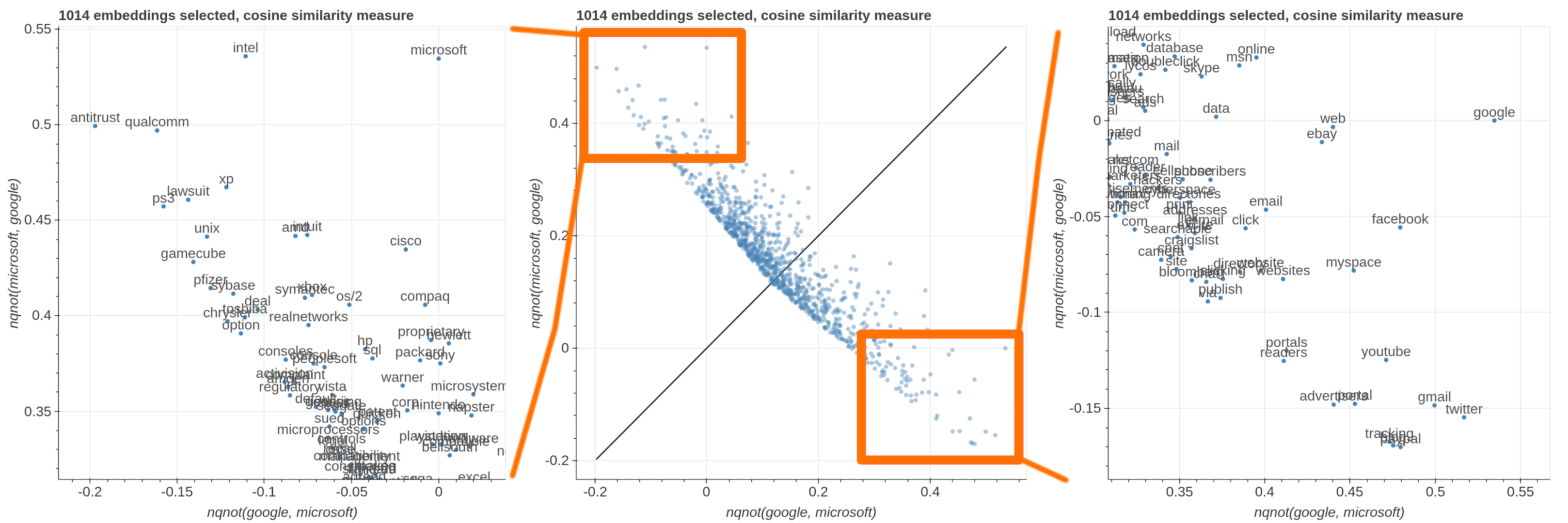}
\caption{Fine-grained comparison of the subspace on the axis $nqnot(google,microsoft)$ and $nqnot(microsoft,google)$ in \textit{Wikipedia}.}
\end{sidewaysfigure*}

\clearpage

\begin{figure*}
\includegraphics[width=\textwidth]{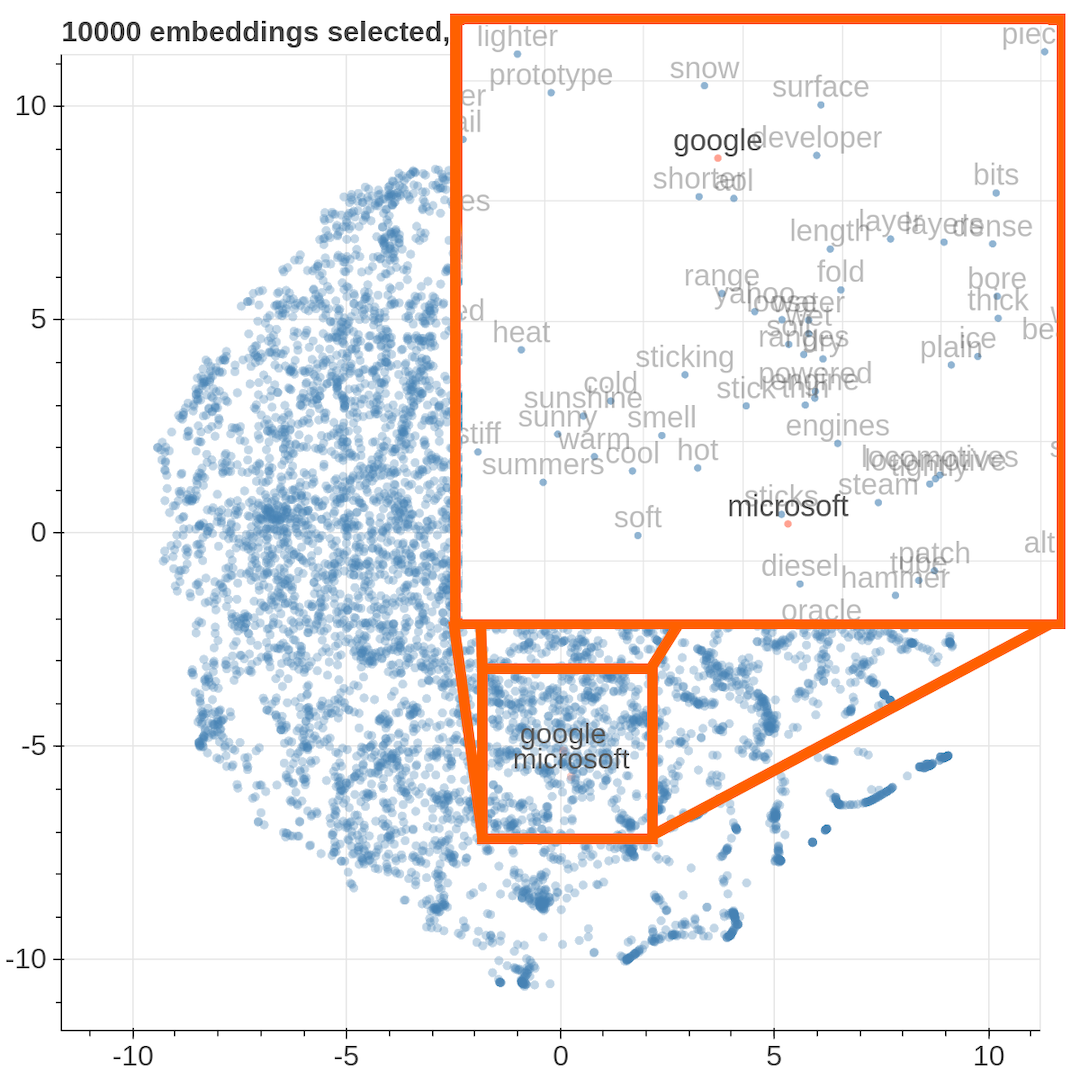}
\caption{t-SNE visualization of \textit{google} and \textit{microsoft} in \textit{Wikipedia}.}
\end{figure*}

\clearpage

\begin{figure*}
\centering
\includegraphics[height=0.9\textheight]{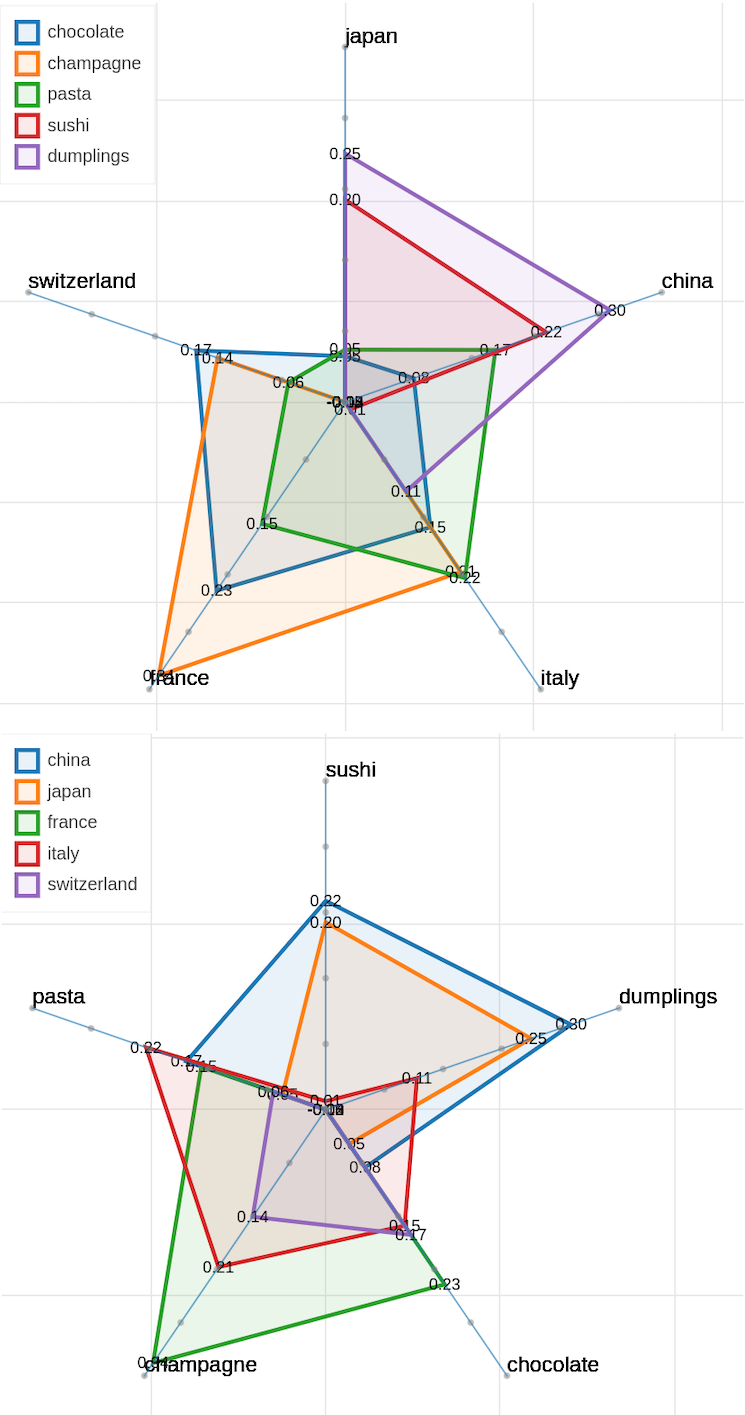}
\caption{Two polar view of countries and foods in \textit{Wikipedia}.}
\end{figure*}

\clearpage

\begin{figure*}
\includegraphics[width=\textwidth]{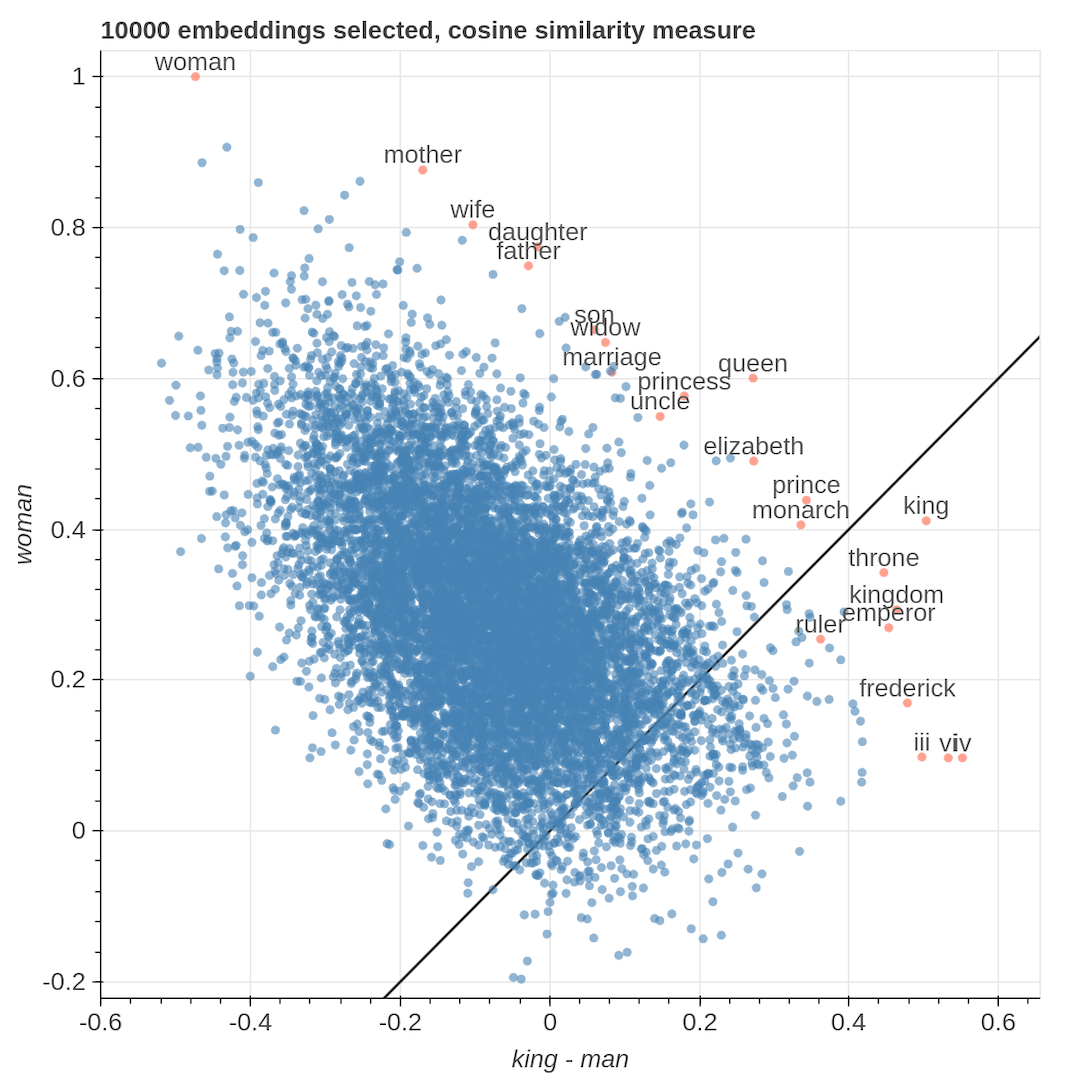}
\caption{king-man vs woman}
\end{figure*}

\clearpage

\begin{figure*}
\includegraphics[width=\textwidth]{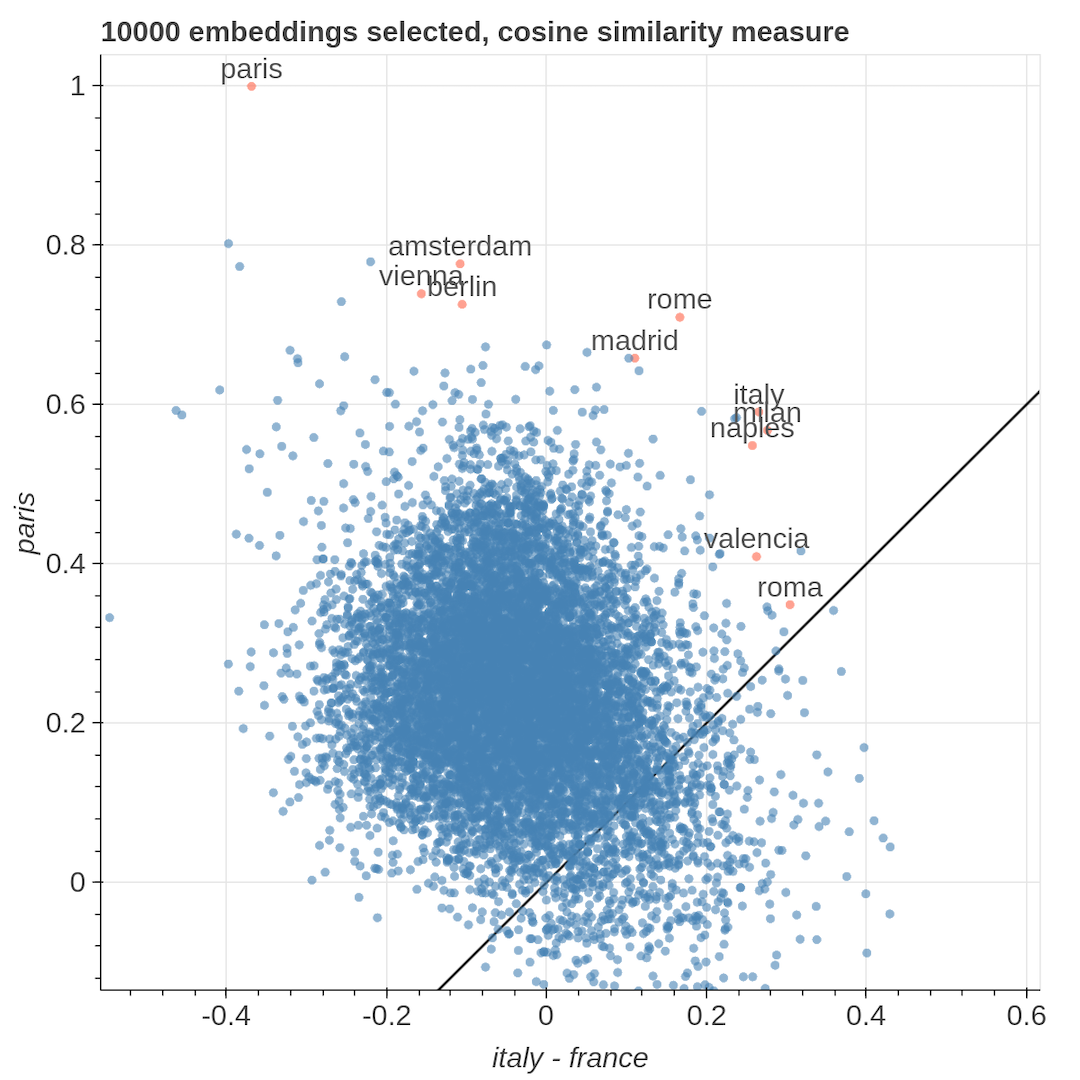}
\caption{italy-france vs paris}
\end{figure*}

\clearpage

\begin{figure*}
\includegraphics[width=\textwidth]{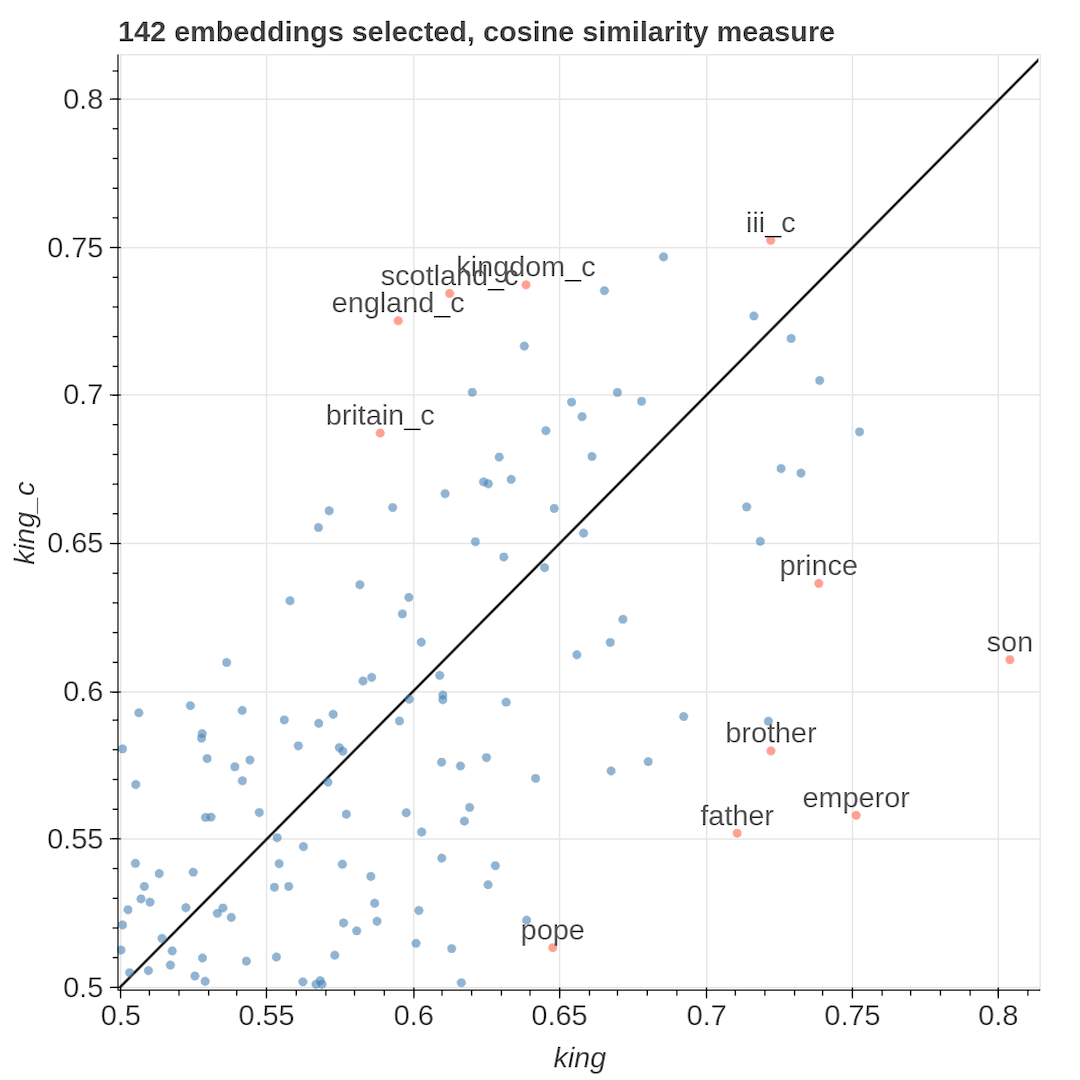}
\caption{paradisgmatic vs syntagmatic}
\end{figure*}

\clearpage

\begin{figure*}
\includegraphics[width=\textwidth]{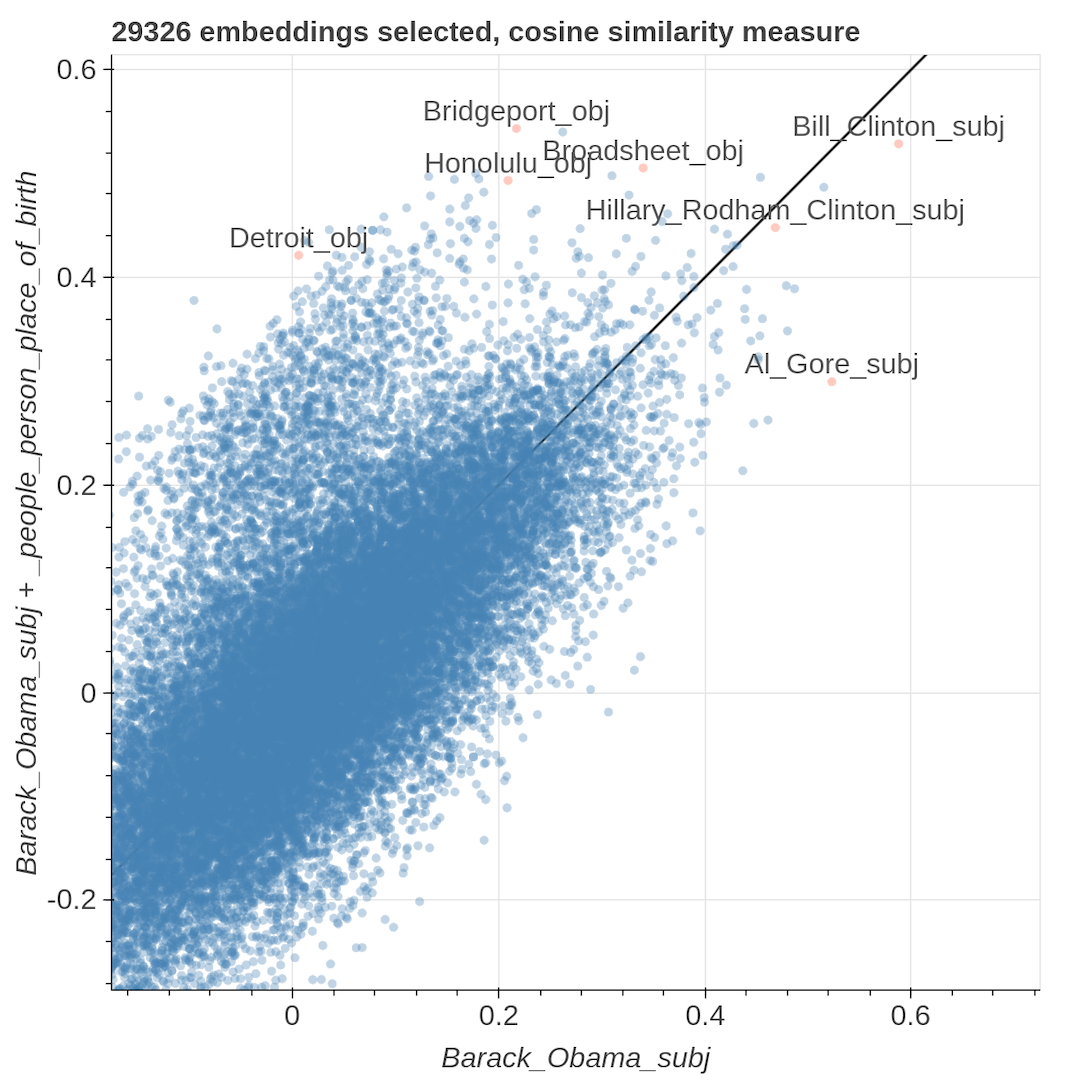}
\caption{knowledge bases}
\end{figure*}

\clearpage

\end{document}